\documentclass[sigconf]{acmart}
\AtBeginDocument{%
  }

\setcopyright{none}
\renewcommand\footnotetextcopyrightpermission[1]{}
\usepackage{booktabs}
\usepackage{multirow}
\usepackage{geometry}
\usepackage{subcaption}
\usepackage{graphicx}
\usepackage{pifont}
\usepackage{xcolor}    % 用于颜色
\usepackage{array}
\usepackage{enumitem}
\usepackage{microtype}
\usepackage{balance}
\usepackage{dblfloatfix}

\providecommand{\tightlist}{%
  \setlength{\itemsep}{0pt}\setlength{\parskip}{0pt}}

\newcommand{\cmark}{\ding{51}} % 对勾
\newcommand{\xmark}{\ding{55}} % 叉号

\begin{document}

\title{CloudCons: A Comprehensive End-to-End Benchmark for Cloud Resource Consolidation}

\author{Xiaobin Zhang}
\affiliation{%
  \institution{Zhejiang University}
  \city{Hangzhou}
  \country{China}
}
\email{zhangxiaobin123@zju.edu.cn}

\author{Lefei Shen}
\affiliation{%
  \institution{Zhejiang University}
  \city{Hangzhou}
  \country{China}
}
\email{lefeishen@zju.edu.cn}

\author{Mouxiang Chen}
\affiliation{%
  \institution{Zhejiang University}
  \city{Hangzhou}
  \country{China}
}
\email{chenmx@zju.edu.cn}

\author{Zhuo Li}
\affiliation{%
  \institution{State Street Technology (Zhejiang) Ltd.}
  \city{Hangzhou}
  \country{China}
}
\email{lizhuo@zju.edu.cn}

\author{Hongkai Li}
\affiliation{%
  \institution{Zhejiang University}
  \city{Hangzhou}
  \country{China}
}
\email{22521276@zju.edu.cn}

\author{Han Fu}
\affiliation{
  \institution{Richoo AI}
  \city{Hangzhou}
  \country{China}
}
\affiliation{
  \institution{Zhejiang University}
  \city{Hangzhou}
  \country{China}
}
\email{fuhan@richoo.ai}

\author{Jianling Sun}
\affiliation{%
  \institution{Zhejiang University}
  \city{Hangzhou}
  \country{China}
}
\email{sunjl@zju.edu.cn}

\author{Xiaoxue Ren}
\authornote{Corresponding authors.}
\affiliation{%
  \institution{Zhejiang University}
  \city{Hangzhou}
  \country{China}
}
\affiliation{%
  \institution{Hangzhou High-Tech Zone (Binjiang) Institute of Blockchain and Data Security}
  \city{Hangzhou}
  \country{China}
}
\email{xxren@zju.edu.cn}

\author{Chenghao Liu}
\authornotemark[1]
\authornote{Work was done before joining Datadog.}
\affiliation{%
  \institution{Datadog AI Research}
  \city{Paris}
  \country{France}
}
\email{twinsken@gmail.com}

\renewcommand{\shortauthors}{Xiaobin Zhang et al.}

\begin{abstract}
Driven by conservative over-provisioning to guarantee service reliability, resource utilization in cloud data centers remains at low levels. To mitigate this, the forecast-then-optimize paradigm has emerged to optimize consolidation by anticipating future demands. While emerging time series foundation models promise to enhance this paradigm through zero-shot generalization, existing benchmarks focus solely on prediction error metrics. The actual decision utility of these advanced models remains unverified, rendering their practical value for downstream tasks uncertain. To bridge this gap, we propose CloudCons, a comprehensive end-to-end benchmark designed to evaluate forecasting models within the specific context of cloud resource consolidation. We build high-quality datasets that cover diverse workloads from Huawei Cloud, Microsoft Azure, and Google Borg, capturing distinct service characteristics ranging from synchronized diurnal rhythms to stochastic, pulse-like bursts and high-frequency noise. We conduct an extensive evaluation of statistical, deep learning, and foundation models. Our experiments reveal a pivotal finding: while foundation models demonstrate superior zero-shot forecasting accuracy, this advantage does not inherently translate into better decision utility. Of practical significance, we systematically analyze how the selection of predictive quantiles acts as a critical lever. We provide actionable guidelines for calibrating these selections to balance the trade-off between resource efficiency and service reliability, offering vital insights for real-world deployment decisions.
\vspace{-0.5em}
\end{abstract}





\maketitle

\section{Introduction} 
%在数字化转型浪潮中，现代软件正经历着快速演变。以云计算、软件即服务（SaaS）、微服务和开发运维（DevOps）为代表的新兴基础设施、技术和设计模式，显著提高了软件开发效率。然而，它们也催生了规模和复杂性前所未有的 IT 系统。这种复杂性的急剧增加使得传统的 IT 管理方法难以为继，因此迫切需要一种更智能、更自动化的运维模式。  说一下Understanding and Predicting Workloads对于Resource Management的重要性。这种重要性会越来越显著，因此对全面且实用的模型评估的需求也在增加。说明forecasting对allocation的重要作用。foundation model 在AIOPS领域有很多有优势

%motivation（现有benchmark有哪些问题）：1.数据多样性严重匮乏（数据集当作benchmark）：时间粒度单一，不同时间粒度下季节性、趋势、波动性会呈现较大差异，这种单一性与现实应用场景严重脱节——例如，异常检测任务通常依赖高频数据，而自动伸缩（Autoscaling）等决策任务则更关注低频趋势；数据来源同质：即便在同一领域，来自不同底层系统的数据也常伴有显著的分布偏移乃至异构性。现有基准未能整合多源数据，导致模型评估的覆盖面不足，无法全面衡量其泛化能力；现有基准中的数据模式往往是单一且高度集中的。这使得评估结果过于乐观，模型可能只是“过拟合”于特定模式，其在处理真实世界中复杂、多变数据时的脆弱性被严重掩盖 2.评估任务与应用场景脱节，现有基准的任务范围过于狭窄，过度聚焦于预测误差指标的数值高低。这种评估方式忽视了模型在特定场景下解决下游实际问题的真实效能。

%说明评估任务（资源分配）的难度，如何解决它

%contribution: 1.数据收集、清洗的框架；收集并清洗了一个新的数据集，多时间尺度、源自多底层系统、特征分布广 2.包含时序预测与资源分配任务的双任务基准，评估多种时序预测模型、资源分配算法，面向决策的评估指标；端到端的直面实际任务的评估框架 3.详细的实验分析

\begin{table*}[t]
    \centering
    \caption{Comparison of CloudCons with existing cloud benchmarks.}
    \vspace{-0.8 em}
    \label{tab:benchmark_comparison}
    \small
    % 保持之前的修改: 减小列间距 (4pt) 以防止溢出
    \setlength{\tabcolsep}{4pt} 
    
    % 如果表格仍然溢出，可以取消下面这行注释，强制缩放表格到文本宽度
    % \resizebox{\textwidth}{!}{
    
    \begin{tabular}{lcccccc}
        \toprule
        % 表头部分
        \multirow{2}{*}{\textbf{Benchmark}} & 
        \multirow{2}{*}{\textbf{Multi-source}} & 
        \multicolumn{3}{c}{\textbf{Evaluated Models}} & 
        \multirow{2}{*}{\textbf{Quantile Analysis}} & 
        \multirow{2}{*}{\textbf{Tasks}} \\
        \cmidrule(lr){3-5} 
         & & \textbf{Statistical} & \textbf{Deep Learning} & \textbf{Foundation} & & \\
        \midrule
        
        % 第一行: Toner et al. (Huawei Benchmark)
        Toner et al. (2025) \cite{huaweibench} & \textcolor{gray}{\xmark} & \cmark & \textcolor{gray}{\xmark} & \cmark & \textcolor{gray}{\xmark} & Forecasting \\ 
        
        % 第二行: Boom
        \textsc{BOOM \cite{toto}}       & \textcolor{gray}{\xmark} & \cmark & \textcolor{gray}{\xmark} & \cmark & \textcolor{gray}{\xmark} & Forecasting \\ 
        
        % 第三行: CloudOps
        \textsc{CloudOps \cite{cloudops}}  & \cmark & \cmark & \cmark & \textcolor{gray}{\xmark} & \textcolor{gray}{\xmark} & Forecasting \\ 
        
        % 第四行: CloudCons (本工作)
        \textbf{CloudCons}       & \cmark & \cmark & \cmark & \cmark & \cmark & Forecasting \& Consolidation \\ 
        \bottomrule
    \end{tabular}
    
    % } % 如果使用了 resizebox，请取消该行注释以闭合花括号
\end{table*}

Ensuring high service reliability without compromising resource efficiency remains a fundamental challenge in cloud data centers. Currently, driven by conservative over-provisioning to guarantee strict Service Level Agreements (SLAs), global average CPU utilization remains at low levels of 15\%-20\% \cite{hsieh2020utilization,azure2019,borg,huawei2025}. To navigate this trade-off, resource consolidation leverages virtual machine (VM) migration as a dual-purpose mechanism: it not only relieves resource hotspots to safeguard reliability but also aggregates fragmented workloads onto fewer physical machines (PMs) to optimize computational density \cite{con-survey1,vm-review}. Since traditional reactive strategies utilize static thresholds to detect anomalies, they trigger consolidation only after an undesirable state is reached. Consequently, this approach fails to preempt service violations during load spikes, whereas during low-utilization periods, it prolongs resource wastage by delaying the consolidation of fragmented resources. Furthermore, their inability to anticipate future trends often leads to unnecessary migrations driven by transient workload jitter, rather than actual demand shifts \cite{con-survey1,vm-review}. To mitigate these issues, the forecast-then-optimize paradigm \cite{investigation} has emerged as a superior alternative. Specifically, this approach first generates future demands via forecasting models, which then serve as input information for optimization algorithms to produce the final consolidation decisions. By guiding such proactive decisions, this paradigm minimizes the frequency of costly reactive migrations and enhances overall service performance.

The efficacy of the forecast-then-optimize paradigm fundamentally depends on performance in the forecasting stage. This task has traditionally been addressed by statistical methods and deep learning models. However, their practical application is constrained: statistical models struggle with complex non-linearities, while deep learning models lack the zero-shot generalization required for dynamic environments \cite{gift}. Given the massive scale and frequent concept drifts of cloud workloads, the resulting need for dataset-specific retraining imposes prohibitive computational and latency overheads, hindering timely adaptation to unseen behaviors \cite{darlow2024dam}. In contrast, emerging time series foundation models (e.g., Moirai, Chronos, TimesFM) offer a breakthrough \cite{moirai2,chronos2,timesfm25}. Leveraging pre-training on massive, heterogeneous datasets, these models capture universal temporal patterns that enable zero-shot generalization across diverse domains, including volatile cloud workloads \cite{gift,toto}. While these models theoretically promise enhanced cloud resource management, their practical efficacy in the context of resource consolidation remains unverified. A critical question persists: do superior forecasting capabilities actually translate into better decision utility? Answering this is essential, as empirical evidence suggests that high forecasting accuracy does not guarantee optimal consolidation decisions due to the misalignment between prediction metrics and decision-making objectives \cite{decision1,decision2}.

Validating this relationship is impeded by the absence of a standardized benchmark tailored for this scenario. As summarized in Table \ref{tab:benchmark_comparison}, existing cloud benchmarks, such as BOOM, CloudOps, and the work by Toner et al. \cite{toto,cloudops,huaweibench}, exhibit significant limitations. \textbf{First,} regarding task alignment, these benchmarks focus on forecasting accuracy, failing to bridge the critical gap between prediction error and downstream decision quality. They cannot verify whether the emerging zero-shot capabilities of foundation models actually translate into tangible improvements in resource consolidation. \textbf{Second,} concerning data diversity, BOOM and Toner et al. rely on single-platform sources with homogeneous workloads, resulting in skewed and conflicting evaluations about the superiority of foundation models versus baselines. Despite incorporating multi-source data, CloudOps lacks workload characterization. Consequently, it remains unclear whether performance gains stem from model architecture or specific data properties, failing to disentangle model robustness across varying environments. \textbf{Third,} in terms of model coverage, they are insufficient: CloudOps overlooks foundation models, while the others lack comparisons with deep learning baselines, which are essential for verifying the actual advancement of foundation models. 

To address these limitations, we propose \textbf{Cloud Consolidation (CloudCons\footnote{CloudCons is available at https://github.com/dmwyd/CloudCons.}),} an end-to-end benchmark designed for cloud resource consolidation. We integrate heterogeneous real-world workload traces characterized by diverse patterns and conduct a comprehensive evaluation covering statistical, deep learning, and foundation models alongside various optimization methods. Furthermore, we establish an evaluation suite spanning five key dimensions, including prediction error, resource efficiency, load balance, service reliability, and uncertainty quantification, to bridge the gap between forecasting accuracy and decision utility. The core contributions of this work are summarized as follows:

\begin{itemize}[leftmargin=*,topsep=0pt, partopsep=0pt, parsep=0pt, itemsep=0pt] \tightlist 
    \item \textbf{Multi-Cloud Datasets and End-to-End Framework:} We construct high-quality datasets with diverse workload characteristics through a standardized processing pipeline. We further implement a framework that simulates the forecast-then-optimize workflow, breaking the limitations of single forecasting tasks.
    \item \textbf{Benchmarking Foundation Models for Decision Utility:} We conduct a systematic evaluation of time series foundation models for cloud resource consolidation. Unlike prior works that limit evaluation to prediction error, we assess the practical utility of these models in driving consolidation decisions, establishing a new baseline for decision-centric assessment. 
    \item \textbf{In-depth Experimental Analysis and Insights:} We reveal a pivotal finding: although foundation models exhibit superior forecasting accuracy, this advantage does not inherently translate into improvements in downstream decision utility. Furthermore, recognizing that modern models provide probabilistic outputs beyond simple point forecasts, we demonstrate how predictive quantile selection serves as a vital lever to balance the complex trade-off between resource efficiency and service reliability.
 
\end{itemize}

% To address these limitations, we propose TSDec, an end-to-end benchmark designed for resource consolidation. The core contributions of this work are summarized as follows:
% \begin{itemize}[leftmargin=*]
% \tightlist
% \item
% \textbf{Construction of a Unified Data Processing Pipeline:} We establish a standardized framework for data collection and cleaning, releasing a high-quality dataset derived from cloud platforms that covers multiple time scales and possesses rich feature diversity.
% \item 
% \textbf{Establishment of an End-to-End Forecast-then-Optimize Benchmark:} Breaking the limitations of single forecasting tasks, we achieve an end-to-end evaluation from forecasting models to resource consolidation strategies. To this end, we design a decision-oriented evaluation suite that measures model performance from five dimensions: prediction error, resource efficiency, load balance, service reliability, and risk assessment.
% \item 
% \textbf{In-depth Experimental Analysis and Insights:} We unveil a pivotal finding: forecasting accuracy does not inherently guarantee decision utility. Beyond this discovery, we systematically evaluate the impact of predictive quantile selection, demonstrating how it serves as a critical lever to balance resource efficiency against service reliability.
% \end{itemize}

\section{Related Works}

\subsection{Time Series Forecasting Models}
The technological evolution of forecasting methods has progressed from classical linear fitting to the recent emergence of pre-trained foundation models. Classical statistical models (e.g., Seasonal Naive, ARIMA, ETS, and Theta) are characterized by low computational overhead and ease of deployment \cite{hyndman2018forecasting,statsforecast}. However, they often struggle with the non-stationarity and burstiness of cloud workloads. Deep learning architectures, ranging from RNNs \cite{RNN-1} to Transformers \cite{Transformer}, effectively capture complex non-linear patterns but are limited by the requirement of training from scratch on each dataset (e.g., LinearFamily, PatchTST, DeepAR, and TFT) \cite{cloudops,DLinear,PatchTST,DeepAR,TFT}. Current research is increasingly focusing on time series foundation models. These models diverge from traditional pipelines by pre-training on massive datasets, thereby learning robust features that transcend specific domains and sampling frequencies. As representative examples, Moirai 2 \cite{moirai2} employs an encoder-only architecture and has completed pre-training on the "Lotsa" dataset, which contains over 27 billion observations; TOTO \cite{toto} specializes in the observability domain, performing targeted optimization using massive telemetry and internal monitoring data from Datadog; and Chronos 2 \cite{chronos2} recasts time series forecasting as a language modeling task, effectively transferring and leveraging mature transformer architectures from the NLP field. To ensure a comprehensive evaluation, our benchmark also incorporates other emerging foundation models, including TimesFM 2.5 \cite{timesfm25}, Sundial \cite{sundial}, FlowState-9.1M \cite{flow} and Kairos 50M \cite{kairos}.

\subsection{Optimization Methods in Resource Consolidation}
The optimization stage of resource consolidation is typically modeled as an NP-hard bin packing problem subject to strict capacity and assignment constraints \cite{nphard,bin-solve,bin1}. To address this, heuristic algorithms like first fit decreasing (FFD) \cite{FFD-1} and best fit decreasing (BFD) \cite{nphard} have become the standard in both academia and industry due to their computational efficiency. Yet, their greedy nature often traps solutions in local optima, prioritizing execution speed at the expense of global resource efficiency \cite{con-survey1}. Alternatively, meta-heuristic approaches such as ant colony optimization (ACO) \cite{ACO1} and genetic algorithms (GA) \cite{GA1} effectively navigate the solution space to find near-global optima, albeit at a higher computational cost \cite{con-survey1}. For scenarios requiring theoretical optimality, mathematical programming methods like mixed-integer linear programming (MILP) and constraint programming (CP) provide exact solutions and performance upper bounds \cite{CP1,predictive}; however, their exponential time complexity limits scalability in large-scale environments. Finally, while reinforcement learning \cite{rl1,rl2} offers a promising end-to-end alternative (often mapping states directly to actions), it is excluded from this benchmark to maintain a specific focus on the decoupled forecast-then-optimize paradigm.

\begin{figure}[htbp]
    \centering  % 图片全局居中
    \includegraphics[width=0.48\textwidth]{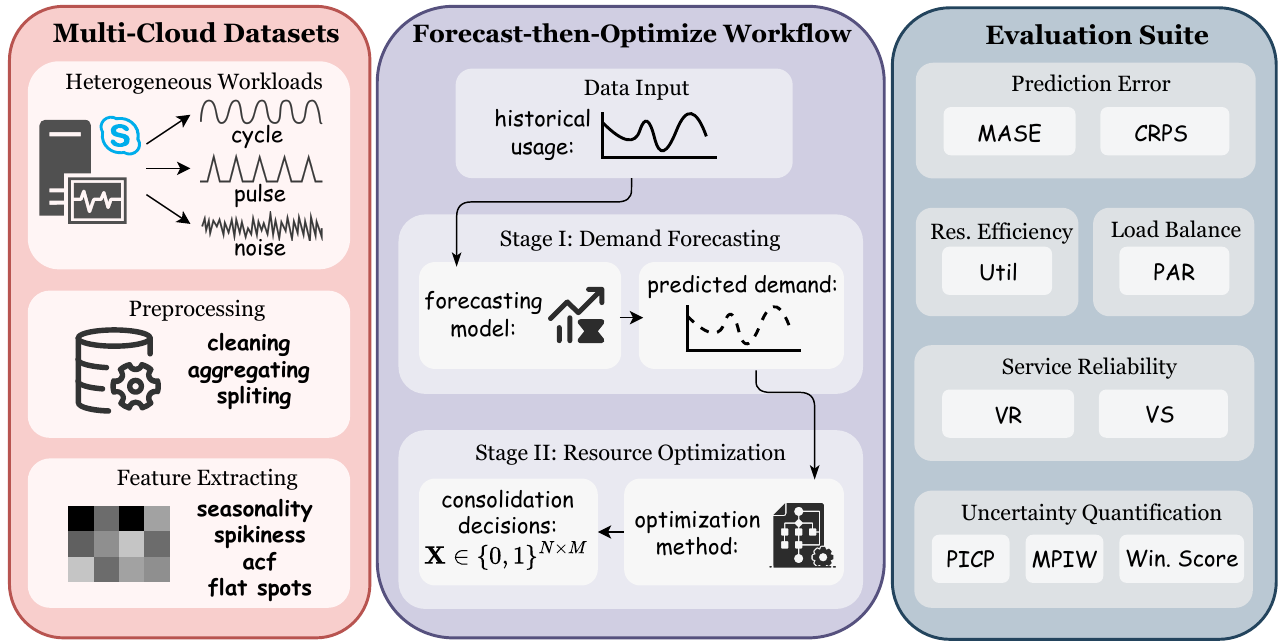}
    \vspace{-1.3em}
    \caption{Overall architecture of CloudCons.}
    \label{fig:bench-frame}
    \vspace{-1 em}
\end{figure}
\section{Benchmark}

This section details the construction of CloudCons. By integrating multi-cloud datasets from the real world, the framework establishes an end-to-end evaluation pipeline designed to analyze the practical utility of forecasting models in complex cloud environments. As illustrated in Figure \ref{fig:bench-frame}, the overall architecture of CloudCons consists of three core modules:

\begin{enumerate}[leftmargin=*]
    \item \textbf{Multi-Cloud Datasets:} This module comprises heterogeneous workload traces collected from Huawei Cloud \cite{huawei2025}, Microsoft Azure \cite{azure2019}, and Google Borg \cite{borg2019}, encompassing various patterns such as periodic, bursty, and noisy loads. 

    \item \textbf{Forecast-then-Optimize Workflow:} As the core execution layer, this module operates in two sequential stages: Stage I generates future demand predictions from historical workloads, which serve as direct inputs for Stage II to derive consolidation decisions via optimization algorithms.

    \item \textbf{Comprehensive Evaluation Suite:} This module moves beyond single error metrics to perform a holistic assessment across five key dimensions: prediction error, resource efficiency, load balance, service reliability, and uncertainty quantification.

\end{enumerate}

\subsection{Datasets}
\label{3-1}
%需要说明数据清洗整理的pipline

\subsubsection{Dataset Construction}
%简单介绍一下数据源，然后说明数据处理过程，包括清洗、聚合、平衡采样。展示一下处理过后的各数据集的统计信息
We construct CloudCons by applying a unified data processing pipeline to real-world workload traces collected from three major cloud providers: Huawei Cloud, Microsoft Azure, and Google Borg. Table \ref{tab:statistics} presents a statistical summary of the curated datasets. Specifically, we integrate diverse data sources: Huawei2025 \cite{huawei2025} captures 31 days of VM and serverless workload traces from a production environment; Azure2019 \cite{azure2019} provides detailed VM workload characteristics, including lifecycle and resource consumption, from Azure's production clusters; and Borg2019 (including Borg2019-d and Borg2019-e) \cite{borg2019} records machine events and instance usage within Google’s compute cells. We implement a rigorous cleaning strategy to ensure data quality. We filter out traces that are excessively short (spanning less than 7 days), contain a high ratio of missing values (exceeding 0.125\%), or lack informational content (i.e., constant values with zero variance). Next, we address data continuity by filling missing values through interpolation. Finally, we aggregate the original high-frequency data into multiple granularities—5-minute, 30-minute, and 1-hour. Through the above steps, we generate four datasets that span multiple time scales and cover diverse workload characteristics.
%对原始数据集进行平衡采样的过程说明
%对资源整合任务特定数据集的进一步清洗说明

\begin{table}[htbp]
\centering
\caption{Key statistics of multi-cloud datasets.}
\vspace{-1 em}
\label{tab:statistics}
\resizebox{\linewidth}{!}{
    \begin{tabular}{l|ccccc}
    \hline
    \multirow{2}{*}{\textbf{Dataset}} & \multirow{2}{*}{\textbf{\# Series}} & \multirow{2}{*}{\textbf{\# Points}} & \multirow{2}{*}{\textbf{\# Targets}} & \multicolumn{2}{c}{\textbf{Length}} \\ \cline{5-6}
     &  &  &  & \textbf{Min} & \textbf{Max} \\ \hline
    Huawei2025  & 174  & 3,352,860  & 2  & 212  & 8,916  \\
    Azure2019  & 10,800  & 104,516,835  & 1  & 188  & 8,628  \\
    Borg2019-d  & 414  & 3,906,696  & 2  & 189  & 8,928  \\
    Borg2019-e  & 618  & 5,901,510  & 2  & 188  & 7,788  \\ \hline
    \end{tabular}
}
\end{table}

% \begin{figure*}[t] 
%     \centering
%     % --- 第一行 ---
%     \begin{subfigure}{0.24\textwidth}
%         \includegraphics[width=\linewidth]{figure/series-huawei2025.jpg}
%     \end{subfigure}
%     \hfill
%     \begin{subfigure}{0.24\textwidth}
%         \includegraphics[width=\linewidth]{figure/series-azure2019.jpg}
%     \end{subfigure}
%     \hfill
%     \begin{subfigure}{0.24\textwidth}
%         \includegraphics[width=\linewidth]{figure/series-borg2019-d.jpg}
%     \end{subfigure}
%     \hfill
%     \begin{subfigure}{0.24\textwidth}
%         \includegraphics[width=\linewidth]{figure/series-borg2019-e.jpg}
%     \end{subfigure}
    
%     \caption{Visualization of CPU utilization time series from multi-cloud datasets. Each subplot displays 30 workload traces randomly sampled from the corresponding dataset.}
%     \label{fig:time_series_plot}
% \end{figure*}

\subsubsection{Dataset Characteristics}
\begin{figure}[htbp]
    \centering  % 图片全局居中
    \includegraphics[width=0.48\textwidth]{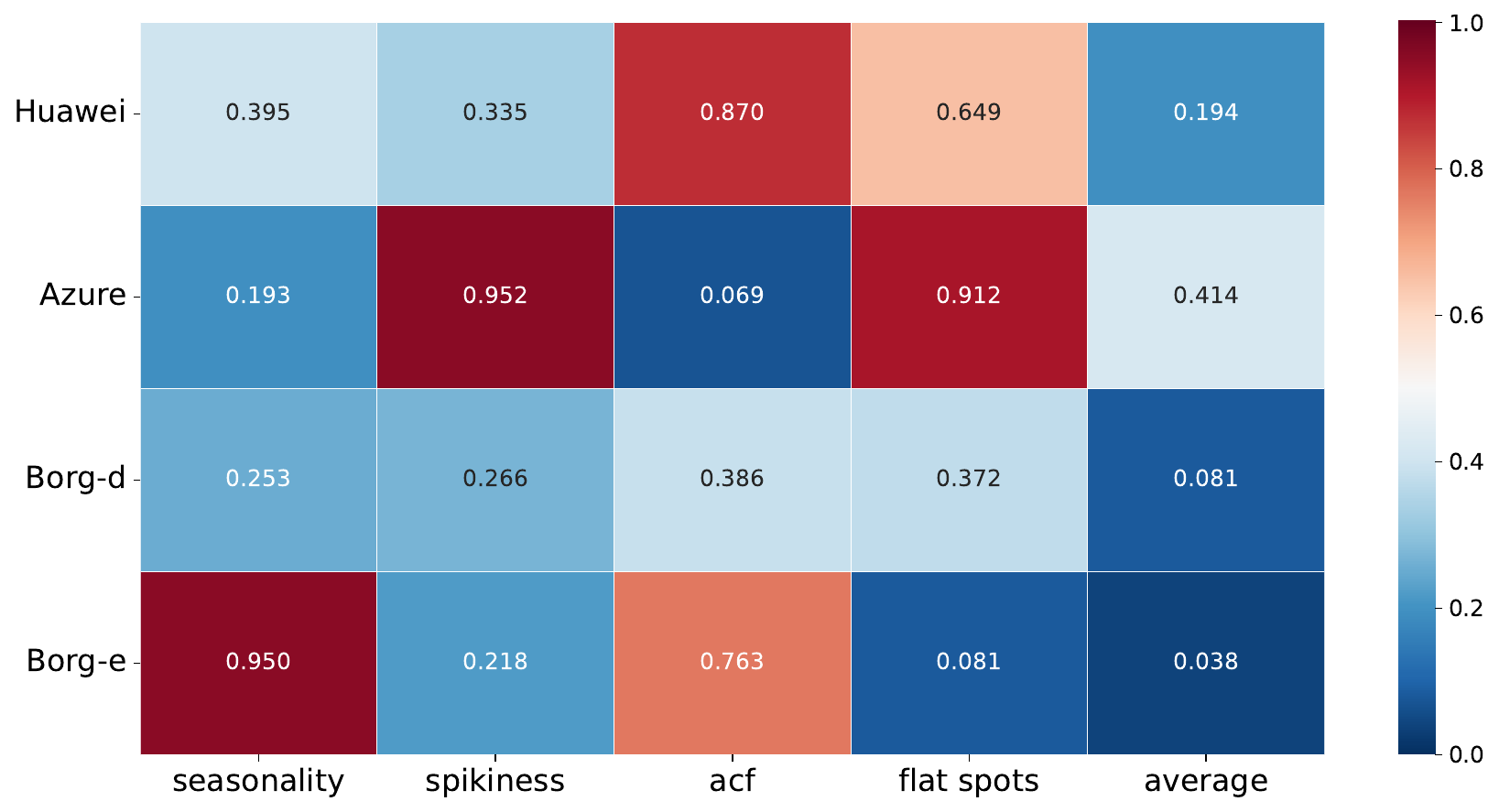}
    \vspace{-1.5 em}
    \caption{Heatmaps depicting mean values of five time series features across different datasets.}
    \label{fig:heatmap}
    \vspace{-0.6 em}
\end{figure}

%分析我们的数据集的时序特征上的特点
%先分析数据集的数据分布在各个时序特征上的特点：
%再结合所有特征看出时序特征分布更加平衡，不至于使得模型“过拟合”于特定模式，评估更公平全面
We incorporate a comprehensive set of time-series features to characterize workload dynamics, specifically seasonality, spikiness, autocorrelation, flat spots level \cite{tsfeature}, and average resource utilization. Seasonality denotes recurrent patterns or cyclic variations observed at fixed time intervals. Spikiness captures the presence of abrupt, high-magnitude, and transient fluctuations or peaks within the data. Autocorrelation quantifies the linear dependence between observations of the same series separated by various time lags. Flat spots identify consecutive intervals where the series exhibits negligible variation or remains static. Finally, within the context of cloud computing, average resource utilization refers to the average amount of resource usage by machines over a period of time. The heatmap in Figure \ref{fig:heatmap} illustrates the mean distribution of features across multi-cloud datasets.

\begin{figure*}[htbp] 
    \centering
    % --- 第一行 ---
    \begin{subfigure}{0.75\textwidth} % 调整宽度，建议 0.6 - 0.9 之间
        \centering
        \includegraphics[width=\linewidth]{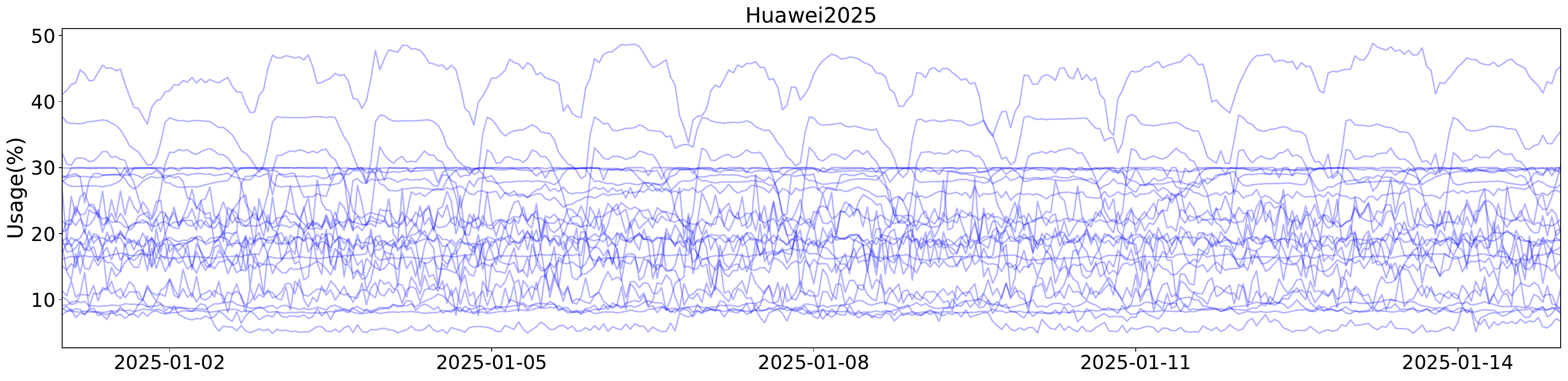}
    \end{subfigure}
    \par\vspace{3pt} % 强制换行并增加垂直间距
    
    % --- 第二行 ---
    \begin{subfigure}{0.75\textwidth}
        \centering
        \includegraphics[width=\linewidth]{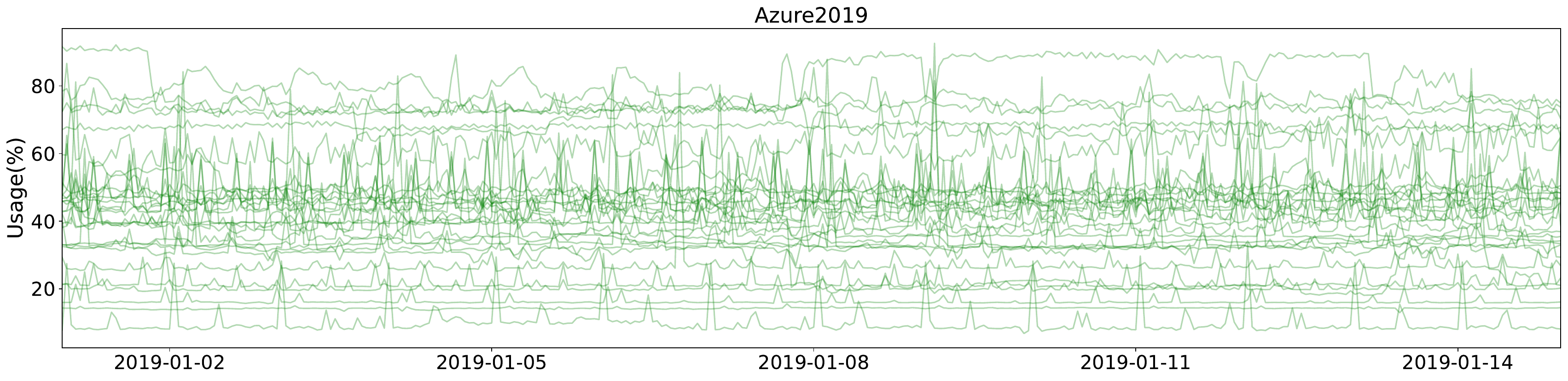}
    \end{subfigure}
    \par\vspace{3pt}
    
    % --- 第三行 ---
    \begin{subfigure}{0.75\textwidth}
        \centering
        \includegraphics[width=\linewidth]{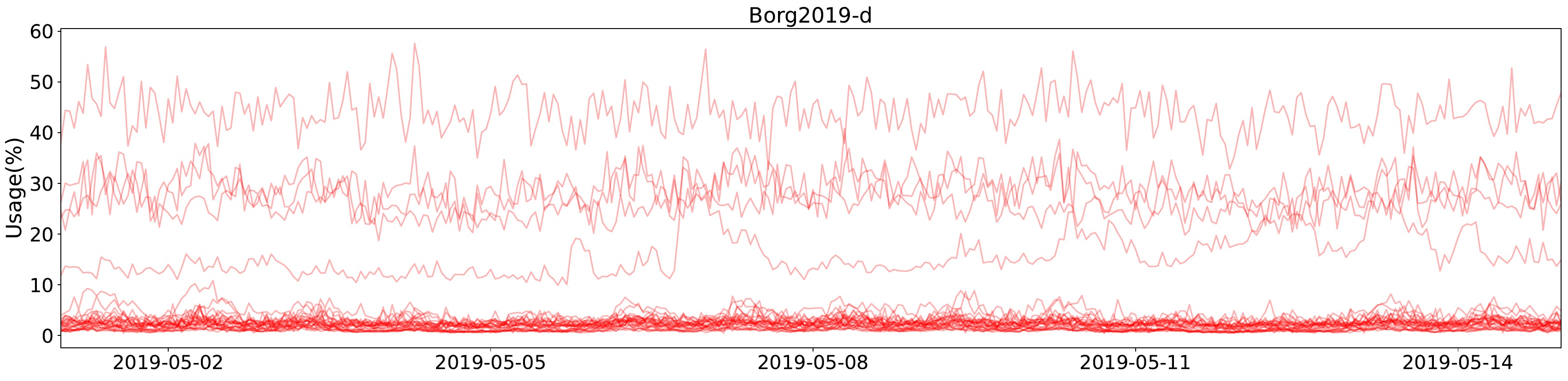}
    \end{subfigure}
    \par\vspace{3pt}
    
    % --- 第四行 ---
    \begin{subfigure}{0.75\textwidth}
        \centering
        \includegraphics[width=\linewidth]{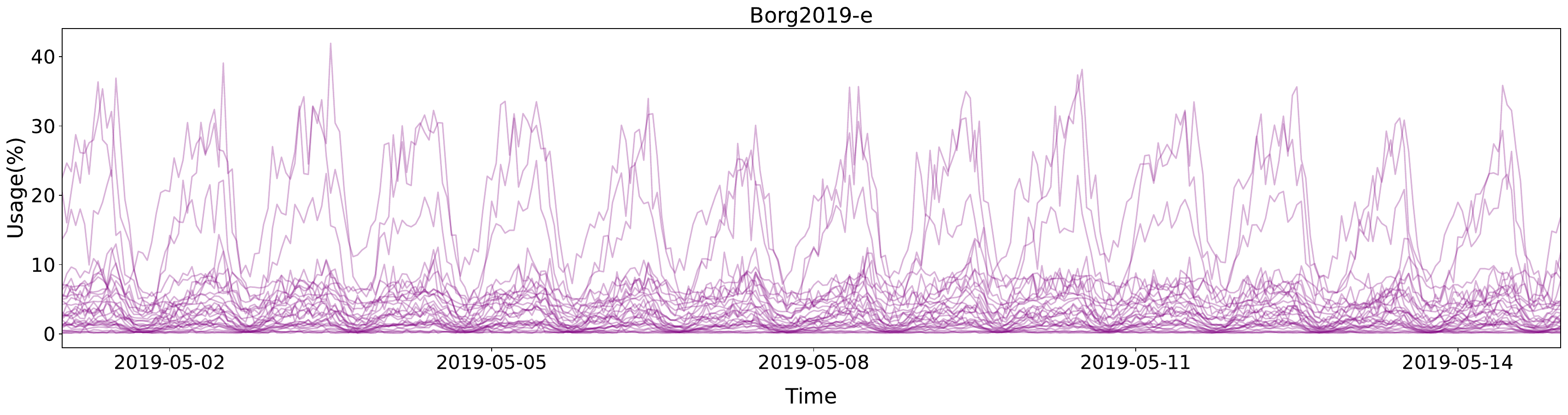}
    \end{subfigure}
    \vspace{-0.8em}
    \caption{Visualization of CPU utilization time series from multi-cloud datasets. Each subplot displays 30 workload traces randomly sampled from the corresponding dataset.}
    \label{fig:time_series_plot}
\vspace{-0.5 em}
\end{figure*}

\textbf{Huawei2025:} The workloads are characterized by temporal dependency and structural heterogeneity. High autocorrelation indicates that the workloads possess significant predictability over short time horizons. Figure \ref{fig:time_series_plot} visualizes examples of traces. Tasks consuming high resources display distinct diurnal patterns, reflecting consistent usage cycles. Conversely, the majority of low-load tasks exhibit stochastic fluctuations. This divergence explains the statistical attenuation of overall seasonal strength despite the high autocorrelation. The mixture of these patterns confirms the complex, non-stationary nature of the environment.

\textbf{Azure2019:} According to Figure \ref{fig:heatmap}, this dataset is characterized by simultaneously high scores in both flat spots level and spikiness. While the high flat spots level indicates that workloads maintain a steady state for the vast majority of time, the elevated spikiness reveals frequent and drastic transient resource demands. These workload traces, visualized in Figure \ref{fig:time_series_plot}, exhibit a distinctive "pulse-like" morphology. Such steep fluctuations, lacking smooth transitions, are emblematic of high-intensity event-driven applications.

\textbf{Borg2019-d and Borg2019-e:} Both datasets originating from Google Borg exhibit an overall characteristic of low resource utilization. Specifically, the vast majority of instances in Borg2019-d hover near a near-zero baseline, accompanied by frequent and chaotic high-frequency jitter. This poses a challenge to the noise robustness of forecasting models. In contrast, Borg2019-e demonstrates exceptional periodic regularity, manifesting a 24-hour diurnal rhythm with "overnight trough and afternoon peak". This highly synchronized pattern likely corresponds to user-facing online services.

\subsection{Forecast-then-Optimize Workflow}
To bridge the gap between forecasting and decision, we establish a unified forecast-then-optimize workflow that transforms historical workload data into actionable consolidation decisions. In resource consolidation, CPU represents the predominant computational bottleneck, and optimizing its utilization efficiency can significantly enhance the performance of cloud systems \cite{luo2021intelligent,luo2021correlation,con-survey1}. Consequently, our experiments adopt CPU load as the representative
metric for resource demand. We consider a set of $N$ VMs, $V = \{v_1, v_2, \ldots, v_N\}$. The evaluation utilizes the historical resource usage sequence ${C}_{i,t} = (c_{i,t-W+1}, \dots, c_{i,t})$, where $W$ is the look-back window and $c_{i,t}$ denotes the resource demand of VM $v_i$ at time step $t$. The entire evaluation process consists of two interconnected stages: first, generating future demands via forecasting models; second, executing consolidation based on the placement plans generated by optimization methods.

\subsubsection{Stage I: Demand Forecasting}
The primary objective of this stage is to transform historical observations into accurate future demand sequences. The forecasting model takes the historical usage sequence ${C}_{i,t}$ as input and generates the predicted demands $\hat{C}_{i,t} = (\hat c_{i,t+1}, \hat c_{i,t+2}, \dots, \hat c_{i,t+H})$ for the forecast horizon $H$. To ensure a comprehensive evaluation, we incorporate a diverse spectrum of models representing the technological evolution of the field:

\begin{itemize}[leftmargin=*]
    \item \textbf{Statistical Models:} Industry-standard methods including AutoARIMA, AutoETS, and AutoTheta \cite{automatic,statsforecast} are utilized for their interpretability and efficiency in capturing linear dependencies.
    \item \textbf{Deep Learning Models:} DeepAR \cite{DeepAR}, TFT \cite{TFT}, DLinear \cite{DLinear}, and PatchTST \cite{PatchTST} are employed to capture complex non-linear temporal patterns through supervised training.
    \item \textbf{Time Series Foundation Models:} We leverage cutting-edge pre-trained models, including Moirai 2 \cite{moirai2}, Chronos 2 \cite{chronos2}, TimesFM 2.5 \cite{timesfm25}, Sundial \cite{sundial}, TOTO \cite{toto}, FlowState-9.1M \cite{flow} and Kairos 50M \cite{kairos}, to evaluate their zero-shot generalization capabilities on heterogeneous cloud workloads.
\end{itemize}

\subsubsection{Stage II: Resource Optimization}
In this stage, the predicted demands $\hat{C}_{i,t}$ serve as inputs to the optimization algorithm to generate a consolidation decision. We formulate this task as a \textbf{dynamic bin packing problem.}

Consider a data center with $N$ VMs to be placed on $M$ available PMs ($P=\{p_1, \dots, p_M\}$), where each PM $p_j$ has a fixed resource capacity $R_j$. We introduce two binary decision variables:
\begin{itemize}[leftmargin=*]
    \item $x_{i,j} \in \{0, 1\}$: An indicator variable that equals 1 if VM $v_i$ is placed on PM $p_j$, and 0 otherwise. We assume the placement decision $x_{i,j}$ remains static throughout the forecast horizon $H$ to minimize migration overhead.
    \item $y_j \in \{0, 1\}$: An indicator variable that equals 1 if PM $p_j$ is active (i.e., hosting at least one VM), and 0 otherwise.
\end{itemize}

The optimization objective is to minimize the total number of active PMs:
\begin{equation}
    \text{Minimize} \quad \sum_{j=1}^{M} y_j
\end{equation}

This problem is subject to the following constraints based on the predicted demands:
\begin{enumerate}[leftmargin=*]
    \item \textbf{Capacity Constraint}: For any active PM $p_j$, the aggregated predicted resource demands of all hosted VMs must not exceed its capacity at any time step $t+\tau$:
    \begin{equation}
        \begin{split}
            \sum_{i=1}^{N} \hat{c}_{i,t+\tau} \cdot x_{i,j} \le R_j \cdot y_j, \\
            \forall j \in \{1, \dots, M\}, \\
            \forall \tau \in \{1, \dots, H\}
        \end{split}
    \end{equation}
    \item \textbf{Assignment Constraint}: Each VM must be assigned to one and only one PM:
    \begin{equation}
        \sum_{j=1}^{M} x_{i,j} = 1, \quad \forall i \in \{1, \dots, N\}
    \end{equation}
\end{enumerate}

To solve it, we employ a spectrum of optimization methods: 
\begin{itemize}[leftmargin=*]
    \item \textbf{Heuristic Algorithms:} We employ first fit decreasing (FFD) and best fit decreasing (BFD), which serve as the core packing logic in modern node auto-provisioning systems like Karpenter \cite{karpenter} and Azure Node Autoprovisioning \cite{azure_nap} due to their low computational latency.
    \item \textbf{Meta-heuristic Algorithms:} We utilize ant colony optimization (ACO), which is a prevalent swarm intelligence algorithm that employs a probabilistic approach inspired by the collective foraging behavior of real ant colonies.
    \item \textbf{Exact Solver:} The Gurobi optimizer \cite{gurobi} provides the theoretical optimal solution by solving the mixed-integer linear programming formulation, serving as the performance upper bound for a given set of forecasts.
\end{itemize}

\subsection{Evaluation Metrics}
Extending beyond traditional time-series benchmarks that focus primarily on prediction errors, we evaluate the methods across five dimensions, integrating standard error metrics with decision-oriented indicators.

\textbf{Prediction Error Metrics:} MASE offers a scale-independent assessment by normalizing the model's MAE against that of a naive baseline. CRPS measures the discrepancy between the predictive CDF $F$ and the ground truth $c_{i,t+\tau}^*$. Their mathematical definitions are formalized as follows:

\begin{equation*}
\text{MASE} = \frac{\frac{1}{H} \sum_{\tau=1}^{H} |c_{i,t+\tau}^* - \hat c_{i,t+\tau}|}{\frac{1}{H} \sum_{\tau=1}^{H} |c_{i,t+\tau}^* - \tilde c_{i,t+\tau}|}
\end{equation*}
where $\tilde c_{i,t+\tau}$ is the naive forecast value.
\begin{equation*}
\text{CRPS($F$,$c_{i,t+\tau}^*$)} =  \int_{0}^{1} 2 \Lambda_\alpha(F^{-1}(\alpha), c_{i,t+\tau}^*) \, d\alpha
\end{equation*}
where $\Lambda_\alpha(q, y) = \max(\alpha(y-q), (1-\alpha)(q-y))$ denotes the quantile loss function.

\textbf{Resource Efficiency:} We define utility ratio (Util) \cite{predictive} as the ratio of the consumed resources to the capacity of an active PM. Higher Util indicates a more aggressive strategy that effectively reduces resource waste. For a PM $p_j$, Util is defined as follows:
    $$
    \text{Util} = \frac{1}{H}  \sum_{\tau=1}^{H} \left( \frac{A_{j,t+\tau}}{R_j} \right)
    $$ 
where $A_{j,t+\tau} = \sum_{i=1}^{N} c_{i,t+\tau}^* \cdot x_{i,j}$ represents the actual aggregated resource usage of $p_j$ at time step $t+\tau$.

\textbf{Load Balance:} Peak-to-Average Ratio (PAR) \cite{par} quantifies the temporal volatility of resource usage on active PMs. Higher PAR indicates that the PM experiences extreme load bursts relative to its average load, implying a heightened risk of performance jitter and thermal throttling. For a PM $p_j$, PAR is defined as follows:
$$\text{PAR} = \frac{\max_{\tau=1}^{H}  A_{j,t+\tau} }{\frac{1}{H} \sum_{\tau=1}^{H} A_{j,t+\tau}}$$

\textbf{Service Reliability:} Violation rate (VR) and violation severity (VS) quantify the frequency and severity of resource contention \cite{predictive}, respectively. VR is calculated as the percentage of time steps where the actual demand of the PM $p_j$ exceeds its fixed capacity $R_j$. VS measures the accumulative magnitude of these resource deficits. Lower values in both metrics indicate that the model effectively anticipates peaks and prevents performance degradation. Note that in the ideal scenario where no resource contentions occur, we define VS as 0.
    $$
    \text{VR} = \frac{\sum_{\tau=1}^{H} \mathbb{I}\left( A_{j,t+\tau} > R_j \right)}{H}
    $$
    $$
    \text{VS} =   \frac{\sum_{\tau=1}^{H}  \max\left( 0, A_{j,t+\tau} - R_j \right)}{R_j \cdot \sum_{\tau=1}^{H} \mathbb{I}\left( A_{j,t+\tau} > R_j \right)}
    $$
    
\textbf{Uncertainty Quantification:} We evaluate the aggregated uncertainty specifically at the PM level. Even if the consolidation decision is based on deterministic point forecasts, the reliability of the resulting system state is inherently probabilistic. Relying solely on service reliability metrics is insufficient to measure potential risks. Therefore, we utilize probabilistic metrics to answer a critical question: Can we trust the safety margins implied by the model? Answering this is challenging because the aggregation of VMs on a single PM introduces complex error superposition. To model this aggregated uncertainty, we employ Monte Carlo simulation \cite{mc}. For a PM $p_j$ hosting a set of VMs, we draw $K$ independent samples from the predictive distribution of each hosted VM and sum the corresponding samples to derive the PM-level predictive distribution. Subsequently, we extract the $(1-\alpha)$ prediction interval $[\hat{L}_{j, t+\tau}^\alpha, \hat{U}_{j, t+\tau}^\alpha]$ from the resulting empirical distribution, where $\alpha$ is set to $0.05$. We evaluate the uncertainty using three metrics \cite{PICP}.

PICP measures the reliability of safety margins, indicating the probability that the ground truth falls within the prediction interval.
$$\text{PICP} = \frac{1}{H}  \sum_{\tau=1}^{H} \mathbb{I}\left( \hat{L}_{j, t+\tau}^\alpha \le A_{j,t+\tau} \le \hat{U}_{j, t+\tau}^\alpha \right)$$

MPIW assesses the sharpness of the aggregated distribution.
$$\text{MPIW} = \frac{1}{H} \sum_{\tau=1}^{H} \left( \hat{U}_{j, t+\tau}^\alpha - \hat{L}_{j, t+\tau}^\alpha \right)$$

Winkler Score balances sharpness and reliability, imposing heavy penalties for out-of-bound violations.
$$\text{Winkler Score} = \frac{1}{H} \sum_{\tau=1}^{H} S_{j, t+\tau}$$
where $S_{j, t+\tau}$ is defined with interval width $\delta_{j, t+\tau} = \hat{U}_{j, t+\tau}^\alpha - \hat{L}_{j, t+\tau}^\alpha$ as:
$$S_{j, t+\tau} = \begin{cases} 
\delta_{j, t+\tau} & \text{if } \hat{L}_{j, t+\tau}^\alpha \le A_{j,t+\tau} \le \hat{U}_{j, t+\tau}^\alpha \\
\delta_{j, t+\tau} + \frac{2}{\alpha}(\hat{L}_{j, t+\tau}^\alpha - A_{j,t+\tau}) & \text{if } A_{j,t+\tau} < \hat{L}_{j, t+\tau}^\alpha \\
\delta_{j, t+\tau} + \frac{2}{\alpha}(A_{j,t+\tau} - \hat{U}_{j, t+\tau}^\alpha) & \text{if } A_{j,t+\tau} > \hat{U}_{j, t+\tau}^\alpha 
\end{cases}$$

\vspace{0.5 em}
\section{Experiments}
This section presents a comprehensive evaluation, targeting three key research questions: \textbf{RQ1:} How do statistical, deep learning, and foundation models perform in terms of forecasting accuracy when faced with diverse and complex cloud workload patterns? \textbf{RQ2:} How does the decision utility of foundation models in a zero-shot setting compare to baselines, and does their forecasting accuracy align with their resource consolidation performance? \textbf{RQ3:} Given the conflict between resource efficiency and service reliability, how does the strategic selection of predictive quantiles serve as a lever to balance this critical trade-off?

To answer these questions, we assess the forecasting models' predictive performance using a non-overlapping rolling window strategy, while evaluating their downstream utility via a sliding window strategy with a stride of 1. For a detailed description, please refer to the implementation details in Appendix \ref{sec:implementation_details}.

\subsection{Predictive Performance Analysis of Forecasting Models (RQ1)}
\label{sec-4.1}

To standardize our performance analysis, we clarify the following prediction protocols: foundation models are evaluated in a zero-shot setting to leverage their generalizable capabilities; deep learning models follow a full-shot approach, trained from scratch on each dataset; and statistical models perform direct inference by fitting parameters to specific historical windows.

Table \ref{tab:prediction-error} presents the forecasting accuracy across four datasets. Foundation models such as Chronos 2 and TimesFM 2.5 demonstrate significant advantages in complex scenarios. Leveraging pre-training on large-scale, cross-domain heterogeneous data, particularly cloud datasets like CloudOps, these models learn transferable temporal knowledge that can be generalized to capture the burstiness of Azure2019 and the heterogeneity of Huawei2025. In contrast, statistical models and simple deep learning baselines like DLinear and DeepAR exhibit suboptimal performance in these scenarios, yielding MASE and CRPS metrics that lag significantly behind the SOTA foundation models. However, this performance disparity narrows on Borg2019-e, which is characterized by strong seasonality and workload homogeneity. In this scenario, PatchTST achieves forecasting accuracy approaching that of the leading foundation models. This suggests that for highly regularized workloads, intrinsic temporal patterns can be adequately learned by traditional architectures, rendering them a viable and potentially more resource-efficient alternative for stable, periodic workloads. 

\noindent \textbf{Summary:} Foundation models demonstrate universal applicability across diverse cloud environments. Despite operating in a zero-shot setting, they outperform statistical baselines and deep learning models that benefited from full-shot training on the target data. They exhibit dominance in complex scenarios while maintaining competitive performance in highly regularized tasks.

\begin{table}[htbp]
\centering
\caption{Comparison of forecasting accuracy. Best results are highlighted in bold.}
\vspace{-0.5 em}
\label{tab:prediction-error}
% 1. 使用 resizebox 强制适应单栏宽度
\resizebox{\columnwidth}{!}{
% 2. 减小列间距，让字体尽可能大一些
\setlength{\tabcolsep}{3pt} 
\begin{tabular}{l cc cc cc cc}
\toprule
\multirow{2}{*}{\textbf{Model}} & \multicolumn{2}{c}{\textbf{Huawei2025}} & \multicolumn{2}{c}{\textbf{Azure2019}} & \multicolumn{2}{c}{\textbf{Borg2019-d}} & \multicolumn{2}{c}{\textbf{Borg2019-e}} \\
\cmidrule(lr){2-3} \cmidrule(lr){4-5} \cmidrule(lr){6-7} \cmidrule(lr){8-9}
 & \textbf{MASE} & \textbf{CRPS} & \textbf{MASE} & \textbf{CRPS} & \textbf{MASE} & \textbf{CRPS} & \textbf{MASE} & \textbf{CRPS} \\ 
\midrule
AutoARIMA & 1.027 & 0.980 & 0.923 & 0.781 & 0.913 & 0.904 & 0.948 & 0.617 \\
AutoETS & 0.987 & 1.165 & 0.981 & 0.944 & 0.896 & 0.976 & 0.987 & 0.660 \\
AutoTheta & 1.121 & 1.184 & 0.918 & 0.885 & 0.947 & 1.069 & 0.943 & 0.687 \\
\midrule
DLinear & 1.766 & 1.380 & 0.957 & 0.604 & 0.859 & 0.928 & 0.977 & 0.697 \\
PatchTST & 1.050 & 0.838 & 0.726 & 0.400 & 0.731 & 0.668 & 0.870 & 0.371 \\
DeepAR & 1.554 & 1.234 & 0.946 & 0.519 & 0.919 & 0.863 & 0.909 & 0.451 \\
TFT & 0.847 & 0.813 & 0.747 & 0.393 & 0.767 & 0.751 & 0.892 & 0.411 \\
\midrule
Moirai 2 & 0.869 & 0.760 & 0.713 & 0.364 & 0.774 & 0.717 & 0.852 & 0.336 \\
Chronos 2 & \textbf{0.752} & \textbf{0.642} & \textbf{0.690} & 0.364 & \textbf{0.702} & \textbf{0.625} & 0.880 & 0.341 \\
TimesFM 2.5 & 0.789 & 0.700 & 0.700 & \textbf{0.354} & 0.726 & 0.665 & \textbf{0.846} & \textbf{0.333} \\
Sundial & 0.801 & 0.773 & 0.721 & 0.397 & 0.734 & 0.739 & 0.865 & 0.373 \\
TOTO & 0.881 & 0.718 & 0.706 & 0.361 & 0.852 & 0.758 & 0.935 & 0.365 \\
FlowState-9.1M & 0.786 & 0.687 & 0.777 & 0.401 & 0.742 & 0.710 & 0.959 & 0.414 \\
Kairos 50m & 1.036 & 0.955 & 0.781 & 0.434 & 0.834 & 0.787 & 0.884 & 0.395 \\
\bottomrule
\end{tabular}
}

\end{table}

\subsection{Decision Utility Analysis in Resource Consolidation (RQ2)}
\label{sec-4.2}

\begin{table*}[htbp]
\centering
\caption{\textbf{Resource Efficiency and Service Reliability.} End-to-end performance metrics (Util, PAR, VR, VS) averaged across all optimization methods. Best results are highlighted in bold.}
\vspace{-1 em}
\label{tab:efficiency-metrics}
\resizebox{\textwidth}{!}{
\begin{tabular}{l|l|ccc cccc ccccccc}
\toprule
\multirow{2}{*}{\textbf{Dataset}} & \multirow{2}{*}{\textbf{Metric}} & \multicolumn{3}{c}{\textbf{Statistical Model}} & \multicolumn{4}{c}{\textbf{Deep Learning Model}} & \multicolumn{7}{c}{\textbf{Foundation Model}} \\
\cmidrule(lr){3-5} \cmidrule(lr){6-9} \cmidrule(lr){10-16}
 & & \textbf{AutoARIMA} & \textbf{AutoETS} & \textbf{AutoTheta} & \textbf{DLinear} & \textbf{PatchTST} & \textbf{DeepAR} & \textbf{TFT} & \textbf{Moirai 2} & \textbf{Chronos 2} & \textbf{TimesFM 2.5} & \textbf{Sundial} & \textbf{TOTO} & \textbf{FlowState-9.1M} & \textbf{Kairos 50m} \\
\midrule
\multirow{4}{*}{\textbf{Huawei2025}} 
 & \textbf{Util} & 0.953 & 0.963 & 0.962 & 0.958 & 0.940 & \textbf{0.969} & \textbf{0.969} & 0.968 & 0.954 & \textbf{0.969} & 0.960 & 0.961 & 0.958 & 0.966 \\
 & \textbf{PAR} & 1.074 & 1.108 & 1.078 & 1.078 & 1.097 & 1.091 & 1.067 & \textbf{1.063} & 1.080 & 1.073 & \textbf{1.063} & 1.083 & 1.078 & 1.071 \\
 & \textbf{VR} & 0.195 & 0.261 & 0.215 & 0.171 & \textbf{0.096} & 0.236 & 0.204 & 0.204 & 0.187 & 0.207 & 0.209 & 0.182 & 0.187 & 0.220 \\
 & \textbf{VS} & 0.014 & 0.018 & 0.015 & 0.016 & \textbf{0.013} & 0.015 & 0.015 & 0.014 & 0.014 & 0.015 & 0.014 & 0.015 & 0.014 & 0.014 \\
\midrule
\multirow{4}{*}{\textbf{Azure2019}} 
 & \textbf{Util} & 0.927 & 0.930 & 0.928 & 0.937 & \textbf{0.973} & \textbf{0.973} & \textbf{0.973} & 0.937 & 0.930 & 0.936 & 0.932 & 0.931 & \textbf{0.973} & \textbf{0.973} \\
 & \textbf{PAR} & 1.100 & 1.096 & 1.093 & 1.108 & 1.080 & 1.080 & 1.115 & 1.094 & 1.105 & 1.121 & 1.108 & 1.112 & \textbf{1.063} & 1.079 \\
 & \textbf{VR} & \textbf{0.110} & 0.122 & 0.116 & 0.135 & 0.223 & 0.287 & 0.474 & 0.162 & 0.134 & 0.158 & 0.152 & 0.159 & 0.199 & 0.240 \\
 & \textbf{VS} & \textbf{0.017} & 0.019 & 0.018 & 0.020 & 0.021 & 0.020 & 0.027 & 0.020 & 0.019 & 0.020 & 0.019 & 0.020 & 0.020 & 0.020 \\
\midrule
\multirow{4}{*}{\textbf{Borg2019-d}} 
 & \textbf{Util} & 0.863 & \textbf{0.918} & 0.868 & 0.892 & 0.849 & 0.825 & 0.883 & 0.898 & 0.864 & 0.884 & 0.901 & 0.885 & 0.864 & 0.908 \\
 & \textbf{PAR} & 1.184 & 1.217 & 1.200 & 1.175 & 1.240 & 1.344 & 1.237 & 1.153 & \textbf{1.135} & 1.169 & 1.147 & 1.163 & 1.170 & 1.158 \\
 & \textbf{VR} & 0.191 & 0.301 & 0.217 & 0.290 & 0.150 & \textbf{0.119} & 0.300 & 0.269 & 0.167 & 0.284 & 0.303 & 0.222 & 0.195 & 0.310 \\
 & \textbf{VS} & 0.061 & 0.087 & 0.069 & 0.077 & 0.057 & 0.058 & 0.078 & 0.069 & \textbf{0.052} & 0.068 & 0.068 & 0.063 & 0.059 & 0.072 \\
\midrule
\multirow{4}{*}{\textbf{Borg2019-e}} 
 & \textbf{Util} & 0.639 & 0.637 & \textbf{0.679} & 0.637 & 0.637 & 0.637 & 0.637 & 0.637 & 0.667 & 0.675 & 0.637 & 0.645 & 0.637 & 0.637 \\
 & \textbf{PAR} & 1.265 & 1.252 & 1.266 & 1.233 & 1.219 & \textbf{1.179} & 1.220 & 1.288 & 1.222 & 1.228 & 1.263 & 1.282 & 1.256 & 1.242 \\
 & \textbf{VR} & 0.092 & \textbf{0.038} & 0.156 & 0.064 & 0.100 & 0.042 & 0.060 & 0.083 & 0.106 & 0.118 & 0.089 & 0.118 & 0.083 & 0.070 \\
 & \textbf{VS} & 0.061 & 0.046 & 0.081 & 0.054 & 0.051 & \textbf{0.044} & 0.052 & 0.069 & 0.063 & 0.072 & 0.063 & 0.074 & 0.061 & 0.059 \\
\bottomrule
\end{tabular}
}
\end{table*}

\begin{table*}[htbp]
\centering
\caption{\textbf{Uncertainty Quantification.} Evaluation using probabilistic metrics (PICP, MPIW, Win. Score) to assess the reliability of safety margins. DLinear is excluded due to the difficulty in generating probabilistic outputs. Best results are highlighted in bold.}
\vspace{-1 em}
\label{tab:risk-metrics}
\resizebox{\textwidth}{!}{
\begin{tabular}{l|l|ccc cccc ccccccc}
\toprule
\multirow{2}{*}{\textbf{Dataset}} & \multirow{2}{*}{\textbf{Metric}} & \multicolumn{3}{c}{\textbf{Statistical Model}} & \multicolumn{4}{c}{\textbf{Deep Learning Model}} & \multicolumn{7}{c}{\textbf{Foundation Model}} \\
\cmidrule(lr){3-5} \cmidrule(lr){6-9} \cmidrule(lr){10-16}
 & & \textbf{AutoARIMA} & \textbf{AutoETS} & \textbf{AutoTheta} & \textbf{DLinear} & \textbf{PatchTST} & \textbf{DeepAR} & \textbf{TFT} & \textbf{Moirai 2} & \textbf{Chronos 2} & \textbf{TimesFM 2.5} & \textbf{Sundial} & \textbf{TOTO} & \textbf{FlowState-9.1M} & \textbf{Kairos 50m} \\
\midrule
\multirow{3}{*}{\textbf{Huawei2025}} 
 & \textbf{PICP} & 0.931 & 0.920 & 0.933 & - & 0.893 & 0.883 & 0.895 & 0.924 & 0.925 & 0.917 & 0.579 & \textbf{0.986} & 0.159 & 0.786 \\
 & \textbf{MPIW} & 0.384 & 0.482 & 0.454 & - & 0.379 & 0.344 & 0.350 & 0.381 & 0.330 & 0.345 & \textbf{0.145} & 1.310 & 0.226 & 0.237 \\
 & \textbf{Win. Score} & 0.650 & 0.778 & 0.644 & - & 0.811 & 0.967 & 0.876 & 0.645 & \textbf{0.640} & 0.702 & 1.246 & 1.348 & 6.418 & 0.890 \\
\midrule
\multirow{3}{*}{\textbf{Azure2019}} 
 & \textbf{PICP} & 0.941 & 0.956 & 0.954 & - & 0.957 & 0.794 & 0.900 & 0.933 & 0.921 & 0.902 & 0.626 & \textbf{0.995} & 0.192 & 0.782 \\
 & \textbf{MPIW} & 0.375 & 0.419 & 0.423 & - & 0.536 & 0.345 & 0.487 & 0.310 & 0.280 & 0.265 & \textbf{0.141} & 1.582 & 0.197 & 0.225 \\
 & \textbf{Win. Score} & 0.576 & 0.532 & 0.539 & - & 0.685 & 0.941 & 0.908 & 0.458 & \textbf{0.454} & 0.523 & 0.915 & 1.593 & 3.545 & 0.783 \\
\midrule
\multirow{3}{*}{\textbf{Borg2019-d}} 
 & \textbf{PICP} & 0.898 & 0.880 & 0.892 & - & 0.879 & 0.613 & \textbf{0.946} & 0.938 & 0.936 & 0.934 & 0.613 & 0.937 & 0.460 & 0.754 \\
 & \textbf{MPIW} & 0.764 & 0.817 & 0.739 & - & 0.377 & \textbf{0.183} & 0.487 & 0.569 & 0.475 & 0.500 & 0.228 & 0.636 & 0.311 & 0.332 \\
 & \textbf{Win. Score} & 1.030 & 1.243 & 1.035 & - & 0.581 & 1.394 & \textbf{0.560} & 0.673 & 0.570 & 0.599 & 1.079 & 0.781 & 2.302 & 0.898 \\
\midrule
\multirow{3}{*}{\textbf{Borg2019-e}} 
 & \textbf{PICP} & 0.865 & 0.866 & 0.778 & - & 0.897 & 0.850 & 0.822 & \textbf{0.931} & 0.881 & 0.880 & 0.466 & 0.930 & 0.200 & 0.746 \\
 & \textbf{MPIW} & 0.238 & 0.243 & 0.230 & - & 0.208 & 0.158 & 0.174 & 0.255 & 0.195 & 0.203 & \textbf{0.085} & 0.363 & 0.138 & 0.141 \\
 & \textbf{Win. Score} & 0.394 & 0.412 & 0.614 & - & 0.287 & \textbf{0.268} & 0.326 & 0.335 & 0.308 & 0.327 & 0.768 & 0.436 & 2.497 & 0.386 \\
\bottomrule
\end{tabular}
}
\vspace{0.3em}
\end{table*}

\begin{figure}[htbp]
    \centering  % 图片全局居中
    \includegraphics[width=0.45\textwidth]{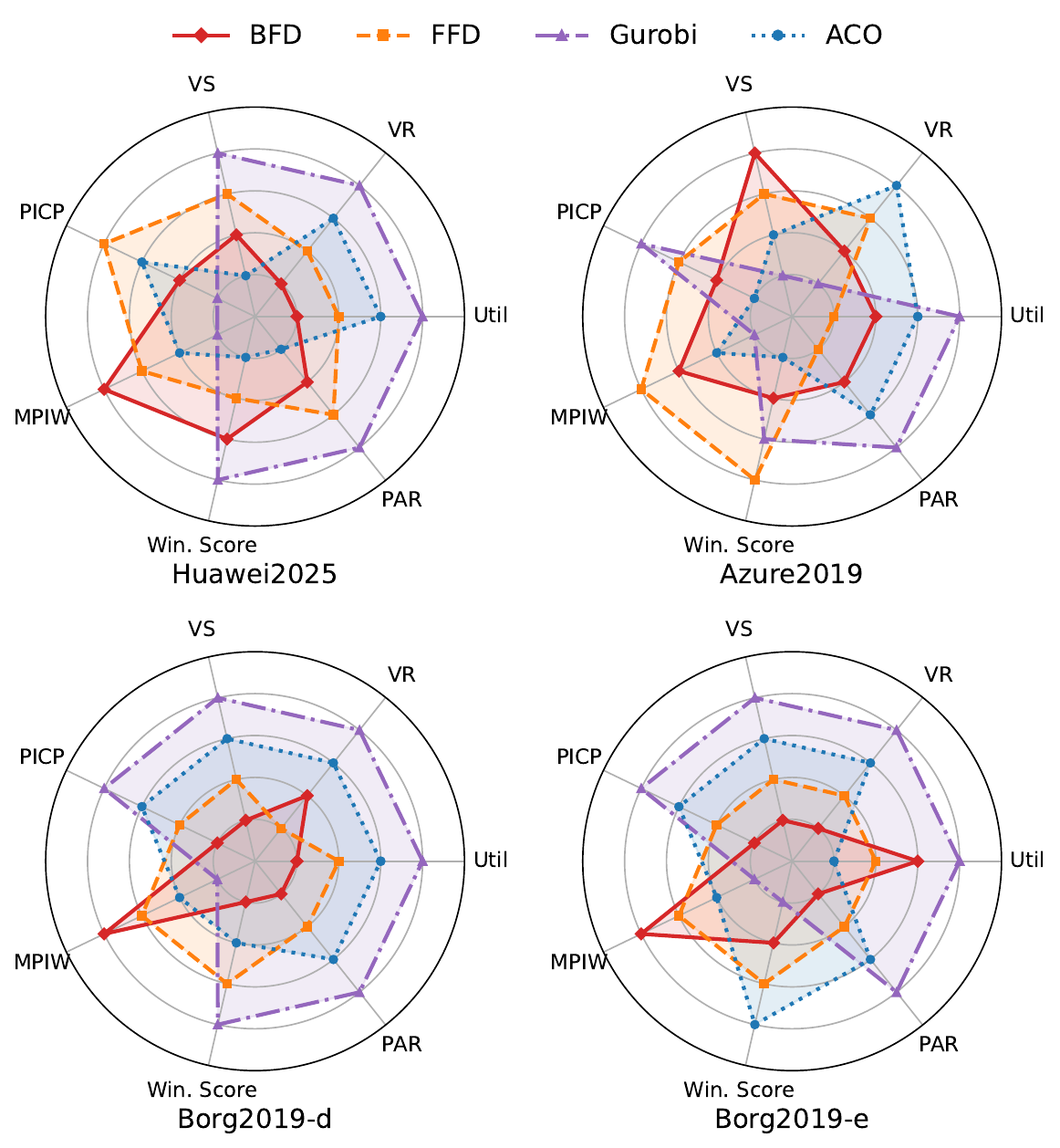}
    \vspace{-0.7 em}
    \caption{Rankings of optimization methods across cloud datasets in resource consolidation.}
    \label{fig:radar}
    \vspace{-0.5 em}
\end{figure}

\noindent \textbf{Ranking Analysis Across Optimization Methods:} 
Before analyzing specific forecasting models, we examine the distinct behavioral patterns of optimization methods as illustrated in radar plots in Figure \ref{fig:radar}. By aggregating metrics across all forecasting models and subsequently ranking the methods, this analysis establishes performance bounds and trade-offs in the decision-making logic. Leveraging global optimization, Gurobi consistently dominates in resource efficiency (Util) and load balancing (PAR). This confirms that upstream forecasts sufficiently capture the temporal phases and multivariate complementarity needed for theoretical optimal packing. However, this high-density consolidation creates a severe risk: the elimination of resource buffers compromises system resilience. This is most evident in bursty environments like Azure2019, where forecast deviations trigger severe degradation in reliability (VR and VS). Conversely, heuristics like BFD sacrifice resource efficiency for robustness; the resulting resource fragmentation acts as an implicit safety margin against uncertainty. Additionally, ACO achieves a balanced performance, ranking second to Gurobi in low-load scenarios (e.g., Borg2019) via its global search mechanism. Given these distinct strategic biases—ranging from Gurobi's aggressive packing to heuristics' conservative buffering—evaluating forecasting models on any single algorithm could yield skewed conclusions. Therefore, the subsequent analysis reports metrics averaged across all optimization methods to mitigate algorithm-specific influence.

\noindent \textbf{Efficiency and Reliability Analysis:} Table \ref{tab:efficiency-metrics} presents the end-to-end performance of all evaluated models focusing on efficiency and reliability metrics. There is a disparity between forecasting accuracy and decision utility. Despite their predictive superiority, foundation models fail to improve resource consolidation, showing no advantage over baselines in violation rates at similar utilization levels. On Huawei2025, the heterogeneity of workloads facilitates efficient resource consolidation, evidenced by low PAR and high utilization rates exceeding 95\% across all models. In this context, PatchTST exhibits a conservative provisioning tendency, trading marginal resource utilization for a twofold improvement in service reliability. On Azure2019, characterized by pulse-like workloads, deep learning models like PatchTST and DeepAR underestimate future peaks, causing the scheduler to over-pack VMs and triggering severe resource contention. In contrast, FlowState-9.1M demonstrates an exceptional balancing capability. By minimizing PAR, it effectively mitigates violation rates while maintaining the highest resource utilization. Dominated by low-utilization tasks and high-frequency noise, Borg2019-d poses severe challenges for predictive modeling. Consequently, we observe a significant decline in resource utilization compared to previous datasets, accompanied by persistently high violation rates. In the Borg2019-e dataset, we observe a utilization bottleneck driven by synchronized diurnal rhythms that restrict resource complementarity. Furthermore, this dataset provides the clearest evidence of the misalignment between forecasting and decision. While most models, ranging from traditional AutoETS, DLinear to advanced Moirai 2, converge to an identical utilization rate of 0.637, their reliability metrics diverge significantly. Baselines like AutoETS and DeepAR achieve both lower VR and VS compared to foundation models. As detailed in our case study (Appendix \ref{case-study}), Moirai 2 and Sundial underestimate the aggregated peaks, misleading the optimizer into risky decisions; conversely, despite exhibiting higher overall prediction errors, baselines like DeepAR effectively avert contention by demonstrating superior fidelity in capturing peak loads. This underscores that for highly regularized workloads, traditional, cost-efficient models offer a more pragmatic solution.

% \begin{figure}[htbp]
%     \centering 
%     \includegraphics[width=0.47\textwidth]{figure/case0.png}
%     \label{fig:case0}
%     \vspace{-1em}
% \end{figure}

\begin{figure*}[htbp]
    \centering 
    \includegraphics[width=0.85\textwidth]{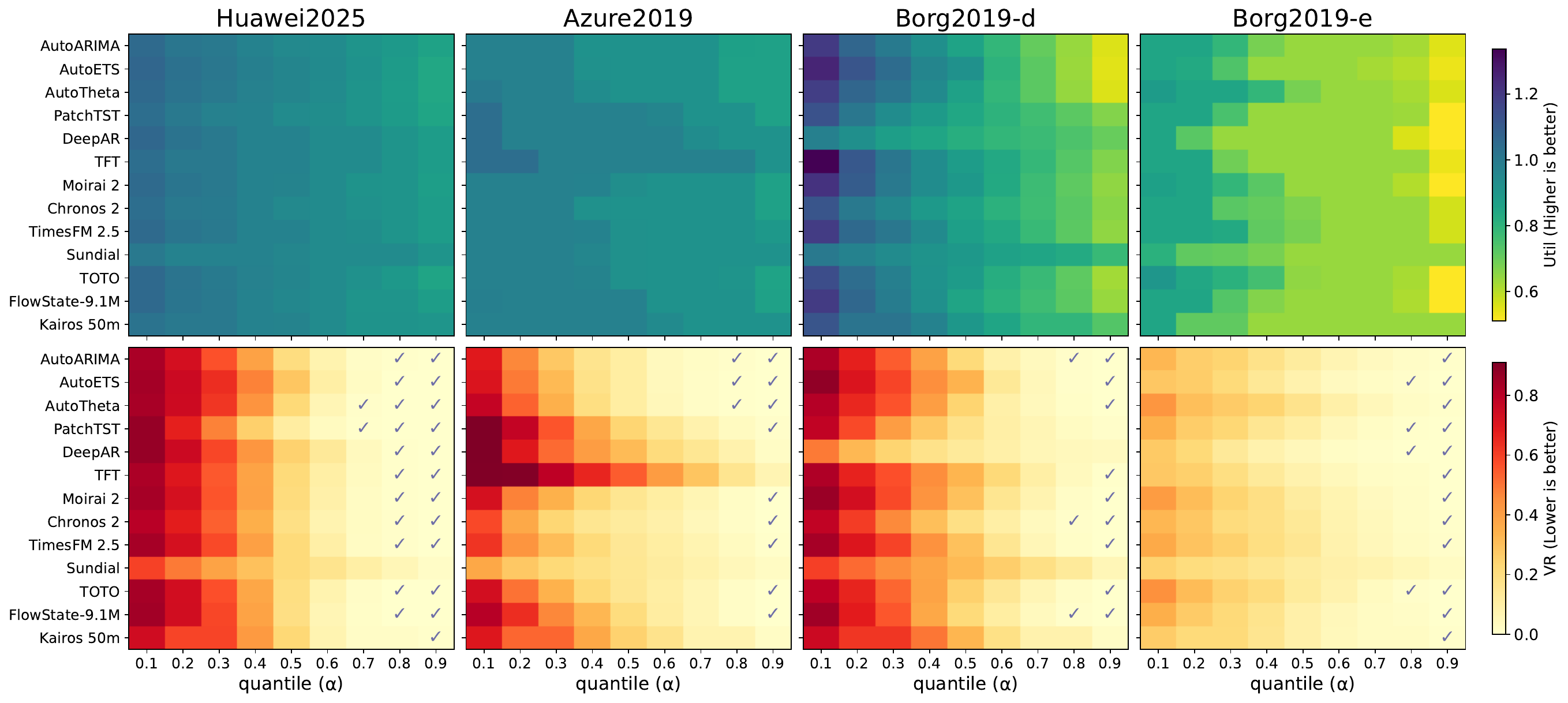}
    \vspace{-1 em}
    \caption{Sensitivity analysis of predictive quantiles ($\alpha$) on Util and VR metrics. DLinear is excluded due to the difficulty in generating probabilistic outputs. Checkmarks ($\checkmark$) denote configurations that achieve the reliability threshold ($VR < 0.01$).}
    \label{fig:quantile}
    \vspace{-1 em}
\end{figure*}

\noindent \textbf{Uncertainty Quantification Analysis:} Table \ref{tab:risk-metrics} presents the uncertainty quantification results. Although optimization methods primarily generate consolidation decisions based on point forecasts, these probabilistic metrics are indispensable for assessing the fidelity of the predictive uncertainty at the PM level.  Chronos 2 demonstrates exceptional probabilistic calibration, attributed to its refined quantile modeling mechanism. By accurately capturing extreme tail behaviors (i.e., the 1st and 99th percentiles), Chronos 2 constructs morphologically precise predictive distributions. This implies that the model provides a reliable probabilistic reference at the PM level, allowing operators to confidently set dynamic safety thresholds or deploy future risk-aware schedulers.
In contrast, FlowState exposes defects related to distributional misalignment. Its critically low PICP reveals that its safety boundaries are deceptive; the systematic bias in the center of its predictive distribution renders its output confidence intervals ineffective. This could mislead operators into erroneous risk assessments. TOTO and statistical baselines (e.g., AutoARIMA) adopt conservative strategies with expanded prediction intervals. For TOTO, this likely stems from its pre-training on massive observability datasets characterized by extreme outliers and heavy-tailed distributions. The model implicitly learns a conservative prior accounting for potential abrupt spikes. While this ensures coverage, it lacks the flexibility to dynamically contract safety margins during stable periods, thereby limiting operational cost reductions.

\noindent \textbf{Summary:} We identify a misalignment between forecasting accuracy and decision utility: superior accuracy does not guarantee better consolidation outcomes. Our experiments show that foundation models struggle to outperform cost-efficient baselines in highly regularized environments, making traditional statistical models the pragmatic choice for predictable workloads. In complex and volatile environments characterized by stochasticity, however, foundation models emerge as a compelling choice. Their primary advantage lies in zero-shot generalization, which enables them to capture intricate patterns and achieve competitive performance without the prohibitive computational costs of retraining or fine-tuning required by deep learning baselines.

\subsection{Sensitivity Analysis of Quantile Selection (RQ3)}
\label{sec-4.3}

To investigate how the selection of predictive quantiles acts as a strategic lever to balance the trade-off between resource efficiency and service reliability, we conduct a sensitivity analysis across $\alpha \in \{0.1, 0.2, \dots, 0.9\}$ using first fit decreasing (FFD) algorithm. FFD is selected to align our benchmark with the core packing logic of widely deployed production schedulers (e.g., Karpenter , Azure NAP ) \cite{karpenter,azure_nap}, ensuring practical relevance while enabling the computational efficiency required for extensive parameter sweeping. As illustrated in the heatmap of Figure \ref{fig:quantile}, a global transition from aggressive to conservative provisioning is observed as $\alpha$ increases, characterized by a simultaneous decline in both Util and VR. A critical threshold for service reliability emerges at $\alpha \ge 0.8$, where a small subset of models first satisfy the stringent reliability requirements (indicated by checkmarks for $VR < 0.01$), suggesting that median-based forecasting ($\alpha = 0.5$) is insufficient to buffer against cloud workload volatility. Specifically, the transition zone between $\alpha = 0.4$ and $0.6$ represents a pivot region where VR exhibits high sensitivity to quantile shifts, whereas the interval $\alpha \in [0.7, 0.9]$ provides a stable regime for high-reliability operations. In complex environments like Huawei2025 and Azure2019, shifting to a higher quantile significantly mitigates resource contention risk with limited impact on overall utilization, making conservative forecasting the most cost-effective choice for these unpredictable scenarios. In the noise-intensive Borg2019-d environment, aggressive low-quantile predictions yield deceptive utilization peaks at the cost of extreme VR, highlighting the severe mismatch risks inherent in stochastic workloads. Conversely, Borg2019-e’s high predictability accommodates aggressive forecasting without compromising reliability, thereby enabling proactive efficiency maximization. However, its consistently lower utilization confirms a ceiling imposed by synchronized diurnal rhythms rather than predictive failure.

\noindent \textbf{Summary and Recommendations:} By leveraging the sensitivity analysis across predictive quantiles, we facilitate a rigorous comparison of violation rates at comparable resource utilization levels. This comparison uncovers distinct performance trade-offs across datasets, explicitly demonstrating that no single model achieves absolute dominance. Based on this, for cost-sensitive applications, we recommend selecting quantiles in the $\alpha \in [0.4, 0.6]$ range. In this sweet spot, the system maintains high resource utilization while significantly reducing potential service violations. Conversely, for mission-critical services, forecasting must be shifted to the high-quantile regime ($\alpha \ge 0.8$) to provide the necessary safety margins.
\section{Conclusion}
In this paper, we propose CloudCons, an end-to-end benchmark designed to bridge the critical gap between time series forecasting accuracy and downstream decision utility in cloud resource consolidation. Integrating diverse workloads, the framework holistically evaluates statistical, deep learning, and foundation models across five key dimensions, ranging from forecasting accuracy to decision utility. The experimental findings reveal that high forecasting accuracy does not inherently guarantee superior system efficiency, highlighting a misalignment that necessitates decision-oriented evaluation metrics rather than traditional error-based benchmarks. Furthermore, the study provides actionable insights through sensitivity analysis on quantile selection, demonstrating that median-based forecasting is suitable for cost-oriented workloads to maximize utilization, whereas selecting quantiles above 80\% is critical for services demanding high reliability to mitigate violation risks, thereby offering practical guidelines for balancing the trade-off between resource efficiency and service reliability.

\begin{acks}
This research is supported by the National Natural Science Foundation of China (No. 62302437) , and Yongjiang Talent Program (No. 2023A-402-G).
\end{acks}

\bibliographystyle{ACM-Reference-Format}
\balance
\bibliography{reference}

\appendix

\section{Implementation Details}
\label{sec:implementation_details}

\begin{table*}[!htbp]
\centering
\caption{Hyperparameter search range for deep learning models.}
\vspace{-0.8 em}
\label{tab:hyperparameters}
\begin{tabular}{l|llll}
\hline
                      & \multicolumn{4}{l}{\textbf{DLinear}}                                                                         \\ \hline
\textbf{Parameters}   & kernel\_size                                 & lr                   &                  &                  \\
\textbf{Search Range} & {[}7,13,25{]}                                & $10^{-4}$                 &                  &                  \\ \hline
                      & \multicolumn{4}{l}{\textbf{PatchTST}}                                                                        \\ \hline
\textbf{Parameters}   & d\_model                                     & num\_encoder\_layers & patch\_len       & lr               \\
\textbf{Search Range} & {[}128,256,512{]}                            & {[}2,3,4{]}          & {[}8,16{]}       & {[}$10^{-5}$,$10^{-2}${]}  \\ \hline
                      & \multicolumn{4}{l}{\textbf{TFT}}                                                                             \\ \hline
\textbf{Parameters}   & num\_heads                                   & hidden\_dim          & lr               &                  \\
\textbf{Search Range} & {[}2,4,8{]}                                  & {[}16,32,64{]}       & {[}$10^{-5}$,$10^{-2}${]}  &                  \\ \hline
                      & \multicolumn{4}{l}{\textbf{DeepAR}}                                                                          \\ \hline
\textbf{Parameters}   & hidden\_size                                 & num\_layers          & lr               &                  \\
\textbf{Search Range} & {[}20,25,30,35,40,45,50,55,60,65,70,75,80{]} & {[}1,2,3,4{]}        & {[}$10^{-5}$,$10^{-1}${]} &                  \\ \hline
\end{tabular}
\end{table*}

\begin{table*}[htbp]
\centering
\caption{\textbf{Resource Efficiency and Service Reliability.} End-to-end performance metrics (Util, PAR, VR, VS) evaluated using the First Fit Decreasing (FFD) algorithm. Best results are highlighted in bold.}
\vspace{-1 em}
\label{tab:efficiency-metrics-ffd}
\resizebox{\textwidth}{!}{
\begin{tabular}{l|l|ccc cccc ccccccc}
\toprule
\multirow{2}{*}{\textbf{Dataset}} & \multirow{2}{*}{\textbf{Metric}} & \multicolumn{3}{c}{\textbf{Statistical Model}} & \multicolumn{4}{c}{\textbf{Deep Learning Model}} & \multicolumn{7}{c}{\textbf{Foundation Model}} \\
\cmidrule(lr){3-5} \cmidrule(lr){6-9} \cmidrule(lr){10-16}
 & & \textbf{AutoARIMA} & \textbf{AutoETS} & \textbf{AutoTheta} & \textbf{DLinear} & \textbf{PatchTST} & \textbf{DeepAR} & \textbf{TFT} & \textbf{Moirai 2} & \textbf{Chronos 2} & \textbf{TimesFM 2.5} & \textbf{Sundial} & \textbf{TOTO} & \textbf{FlowState-9.1M} & \textbf{Kairos 50m} \\
\midrule
\multirow{4}{*}{\textbf{Huawei2025}}
 & \textbf{Util} & 0.948 & 0.960 & 0.960 & 0.957 & 0.939 & \textbf{0.969} & \textbf{0.969} & 0.967 & 0.948 & \textbf{0.969} & 0.956 & 0.960 & 0.953 & 0.966 \\
 & \textbf{PAR} & 1.082 & 1.121 & 1.086 & 1.087 & 1.108 & 1.096 & 1.069 & \textbf{1.065} & 1.092 & 1.079 & 1.074 & 1.093 & 1.088 & 1.074 \\
 & \textbf{VR} & 0.197 & 0.265 & 0.217 & 0.181 & \textbf{0.106} & 0.248 & 0.214 & 0.206 & 0.187 & 0.215 & 0.213 & 0.186 & 0.188 & 0.224 \\
 & \textbf{VS} & 0.014 & 0.018 & 0.015 & 0.016 & \textbf{0.013} & 0.015 & 0.015 & 0.015 & 0.014 & 0.014 & 0.014 & 0.015 & 0.014 & 0.014 \\
\midrule
\multirow{4}{*}{\textbf{Azure2019}}
 & \textbf{Util} & 0.916 & 0.916 & 0.916 & 0.924 & \textbf{0.973} & \textbf{0.973} & \textbf{0.973} & 0.927 & 0.916 & 0.918 & 0.918 & 0.918 & \textbf{0.973} & \textbf{0.973} \\
 & \textbf{PAR} & 1.114 & 1.109 & 1.104 & 1.120 & 1.079 & 1.080 & 1.121 & 1.104 & 1.119 & 1.143 & 1.120 & 1.123 & \textbf{1.057} & 1.080 \\
 & \textbf{VR} & 0.109 & 0.106 & \textbf{0.105} & 0.132 & 0.234 & 0.310 & 0.542 & 0.152 & 0.120 & 0.145 & 0.145 & 0.150 & 0.202 & 0.244 \\
 & \textbf{VS} & \textbf{0.018} & \textbf{0.018} & 0.019 & 0.020 & 0.021 & 0.020 & 0.027 & 0.020 & \textbf{0.018} & 0.020 & \textbf{0.018} & 0.019 & 0.019 & 0.020 \\
\midrule
\multirow{4}{*}{\textbf{Borg2019-d}}
 & \textbf{Util} & 0.854 & \textbf{0.915} & 0.861 & 0.876 & 0.844 & 0.818 & 0.880 & 0.893 & 0.854 & 0.870 & 0.893 & 0.880 & 0.853 & 0.897 \\
 & \textbf{PAR} & 1.206 & 1.224 & 1.220 & 1.177 & 1.248 & 1.357 & 1.242 & 1.156 & \textbf{1.150} & 1.183 & 1.158 & 1.175 & 1.177 & 1.171 \\
 & \textbf{VR} & 0.196 & 0.314 & 0.222 & 0.294 & 0.162 & \textbf{0.124} & 0.327 & 0.280 & 0.170 & 0.293 & 0.309 & 0.229 & 0.201 & 0.312 \\
 & \textbf{VS} & 0.066 & 0.091 & 0.073 & 0.081 & 0.058 & 0.058 & 0.083 & 0.070 & \textbf{0.054} & 0.070 & 0.070 & 0.066 & 0.060 & 0.074 \\
\midrule
\multirow{4}{*}{\textbf{Borg2019-e}}
 & \textbf{Util} & 0.637 & 0.637 & \textbf{0.675} & 0.637 & 0.637 & 0.637 & 0.637 & 0.637 & 0.667 & \textbf{0.675} & 0.637 & 0.645 & 0.637 & 0.637 \\
 & \textbf{PAR} & 1.280 & 1.269 & 1.283 & 1.251 & 1.229 & \textbf{1.201} & 1.235 & 1.307 & 1.233 & 1.242 & 1.275 & 1.300 & 1.278 & 1.254 \\
 & \textbf{VR} & 0.112 & \textbf{0.000} & 0.156 & 0.077 & 0.101 & 0.044 & 0.076 & 0.102 & 0.124 & 0.142 & 0.110 & 0.118 & 0.084 & 0.088 \\
 & \textbf{VS} & 0.059 & \textbf{0.000} & 0.081 & 0.055 & 0.052 & 0.049 & 0.058 & 0.074 & 0.063 & 0.072 & 0.065 & 0.074 & 0.058 & 0.058 \\
\bottomrule
\end{tabular}
}
\vspace{-0.3 em}
\end{table*}

\begin{table*}[htbp]
\centering
\caption{\textbf{Uncertainty Quantification.} Evaluation using probabilistic metrics (PICP, MPIW, Win. Score) to assess the reliability of safety margins under the First Fit Decreasing (FFD) algorithm. DLinear is excluded due to the difficulty in generating probabilistic outputs. Best results are highlighted in bold.}
\vspace{-1 em}
\label{tab:risk-metrics-ffd}
\resizebox{\textwidth}{!}{
\begin{tabular}{l|l|ccc cccc ccccccc}
\toprule
\multirow{2}{*}{\textbf{Dataset}} & \multirow{2}{*}{\textbf{Metric}} & \multicolumn{3}{c}{\textbf{Statistical Model}} & \multicolumn{4}{c}{\textbf{Deep Learning Model}} & \multicolumn{7}{c}{\textbf{Foundation Model}} \\
\cmidrule(lr){3-5} \cmidrule(lr){6-9} \cmidrule(lr){10-16}
 & & \textbf{AutoARIMA} & \textbf{AutoETS} & \textbf{AutoTheta} & \textbf{DLinear} & \textbf{PatchTST} & \textbf{DeepAR} & \textbf{TFT} & \textbf{Moirai 2} & \textbf{Chronos 2} & \textbf{TimesFM 2.5} & \textbf{Sundial} & \textbf{TOTO} & \textbf{FlowState-9.1M} & \textbf{Kairos 50m} \\
\midrule
\multirow{3}{*}{\textbf{Huawei2025}}
 & \textbf{PICP} & 0.930 & 0.916 & 0.930 & - & 0.889 & 0.879 & 0.890 & 0.920 & 0.922 & 0.911 & 0.577 & \textbf{0.984} & 0.195 & 0.782 \\
 & \textbf{MPIW} & 0.377 & 0.476 & 0.448 & - & 0.373 & 0.339 & 0.346 & 0.376 & 0.323 & 0.339 & \textbf{0.143} & 1.308 & 0.222 & 0.234 \\
 & \textbf{Win. Score} & 0.643 & 0.791 & \textbf{0.637} & - & 0.845 & 1.011 & 0.901 & 0.639 & 0.642 & 0.712 & 1.250 & 1.350 & 6.526 & 0.903 \\
\midrule
\multirow{3}{*}{\textbf{Azure2019}}
 & \textbf{PICP} & 0.939 & 0.955 & 0.954 & - & 0.954 & 0.790 & 0.901 & 0.934 & 0.923 & 0.904 & 0.629 & \textbf{0.994} & 0.191 & 0.777 \\
 & \textbf{MPIW} & 0.372 & 0.415 & 0.420 & - & 0.530 & 0.347 & 0.487 & 0.307 & 0.276 & 0.260 & \textbf{0.139} & 1.561 & 0.197 & 0.225 \\
 & \textbf{Win. Score} & 0.585 & 0.532 & 0.532 & - & 0.682 & 0.939 & 0.934 & 0.450 & \textbf{0.448} & 0.509 & 0.902 & 1.572 & 3.547 & 0.807 \\
\midrule
\multirow{3}{*}{\textbf{Borg2019-d}}
 & \textbf{PICP} & 0.880 & 0.864 & 0.876 & - & 0.866 & 0.620 & \textbf{0.940} & 0.926 & 0.923 & 0.920 & 0.593 & 0.924 & 0.454 & 0.739 \\
 & \textbf{MPIW} & 0.749 & 0.809 & 0.727 & - & 0.371 & \textbf{0.182} & 0.481 & 0.561 & 0.465 & 0.486 & 0.223 & 0.629 & 0.305 & 0.327 \\
 & \textbf{Win. Score} & 1.055 & 1.318 & 1.078 & - & 0.591 & 1.398 & \textbf{0.556} & 0.687 & 0.572 & 0.605 & 1.138 & 0.802 & 2.495 & 0.958 \\
\midrule
\multirow{3}{*}{\textbf{Borg2019-e}}
 & \textbf{PICP} & 0.858 & 0.863 & 0.772 & - & 0.893 & 0.857 & 0.814 & \textbf{0.930} & 0.876 & 0.879 & 0.457 & 0.923 & 0.177 & 0.746 \\
 & \textbf{MPIW} & 0.231 & 0.239 & 0.224 & - & 0.204 & 0.155 & 0.172 & 0.248 & 0.191 & 0.199 & \textbf{0.083} & 0.355 & 0.135 & 0.138 \\
 & \textbf{Win. Score} & 0.388 & 0.402 & 0.623 & - & 0.282 & \textbf{0.261} & 0.331 & 0.328 & 0.309 & 0.320 & 0.749 & 0.432 & 2.559 & 0.367 \\
\bottomrule
\end{tabular}
}
\vspace{-1 em}
\end{table*}

\begin{table*}[htbp]
\centering
\caption{\textbf{Resource Efficiency and Service Reliability.} End-to-end performance metrics (Util, PAR, VR, VS) evaluated using the Best Fit Decreasing (BFD) algorithm. Best results are highlighted in bold.}
\label{tab:efficiency-metrics-bfd}
\vspace{-1 em}
\resizebox{\textwidth}{!}{
\begin{tabular}{l|l|ccc cccc ccccccc}
\toprule
\multirow{2}{*}{\textbf{Dataset}} & \multirow{2}{*}{\textbf{Metric}} & \multicolumn{3}{c}{\textbf{Statistical Model}} & \multicolumn{4}{c}{\textbf{Deep Learning Model}} & \multicolumn{7}{c}{\textbf{Foundation Model}} \\
\cmidrule(lr){3-5} \cmidrule(lr){6-9} \cmidrule(lr){10-16}
 & & \textbf{AutoARIMA} & \textbf{AutoETS} & \textbf{AutoTheta} & \textbf{DLinear} & \textbf{PatchTST} & \textbf{DeepAR} & \textbf{TFT} & \textbf{Moirai 2} & \textbf{Chronos 2} & \textbf{TimesFM 2.5} & \textbf{Sundial} & \textbf{TOTO} & \textbf{FlowState-9.1M} & \textbf{Kairos 50m} \\
\midrule
\multirow{4}{*}{\textbf{Huawei2025}}
 & \textbf{Util} & 0.948 & 0.960 & 0.960 & 0.957 & 0.939 & \textbf{0.969} & \textbf{0.969} & 0.967 & 0.948 & \textbf{0.969} & 0.957 & 0.960 & 0.953 & 0.965 \\
 & \textbf{PAR} & 1.082 & 1.124 & 1.086 & 1.087 & 1.109 & 1.096 & 1.069 & \textbf{1.066} & 1.092 & 1.078 & 1.073 & 1.093 & 1.087 & 1.077 \\
 & \textbf{VR} & 0.198 & 0.265 & 0.217 & 0.180 & \textbf{0.106} & 0.248 & 0.214 & 0.205 & 0.187 & 0.215 & 0.214 & 0.187 & 0.189 & 0.223 \\
 & \textbf{VS} & 0.014 & 0.018 & 0.015 & 0.016 & \textbf{0.013} & 0.015 & 0.015 & 0.015 & 0.014 & 0.014 & 0.014 & 0.015 & 0.014 & 0.015 \\
\midrule
\multirow{4}{*}{\textbf{Azure2019}}
 & \textbf{Util} & 0.916 & 0.916 & 0.916 & 0.927 & \textbf{0.973} & \textbf{0.973} & \textbf{0.973} & 0.927 & 0.916 & 0.918 & 0.918 & 0.918 & \textbf{0.973} & \textbf{0.973} \\
 & \textbf{PAR} & 1.115 & 1.109 & 1.103 & 1.118 & 1.079 & 1.079 & 1.122 & 1.104 & 1.120 & 1.143 & 1.122 & 1.124 & \textbf{1.056} & 1.079 \\
 & \textbf{VR} & 0.111 & 0.107 & \textbf{0.106} & 0.134 & 0.234 & 0.310 & 0.543 & 0.154 & 0.123 & 0.149 & 0.147 & 0.152 & 0.204 & 0.247 \\
 & \textbf{VS} & \textbf{0.018} & \textbf{0.018} & 0.019 & 0.020 & 0.021 & 0.019 & 0.027 & 0.020 & \textbf{0.018} & 0.020 & \textbf{0.018} & 0.019 & 0.019 & 0.020 \\
\midrule
\multirow{4}{*}{\textbf{Borg2019-d}}
 & \textbf{Util} & 0.854 & \textbf{0.915} & 0.861 & 0.876 & 0.844 & 0.818 & 0.880 & 0.890 & 0.854 & 0.870 & 0.893 & 0.880 & 0.853 & 0.897 \\
 & \textbf{PAR} & 1.208 & 1.224 & 1.221 & 1.178 & 1.244 & 1.356 & 1.243 & 1.157 & \textbf{1.150} & 1.184 & 1.162 & 1.173 & 1.179 & 1.168 \\
 & \textbf{VR} & 0.196 & 0.314 & 0.220 & 0.292 & 0.163 & \textbf{0.124} & 0.328 & 0.280 & 0.164 & 0.293 & 0.316 & 0.233 & 0.200 & 0.309 \\
 & \textbf{VS} & 0.066 & 0.091 & 0.073 & 0.081 & 0.057 & 0.057 & 0.083 & 0.071 & \textbf{0.055} & 0.070 & 0.070 & 0.068 & 0.060 & 0.073 \\
\midrule
\multirow{4}{*}{\textbf{Borg2019-e}}
 & \textbf{Util} & 0.637 & 0.637 & \textbf{0.675} & 0.637 & 0.637 & 0.637 & 0.637 & 0.637 & 0.667 & \textbf{0.675} & 0.637 & 0.645 & 0.637 & 0.637 \\
 & \textbf{PAR} & 1.280 & 1.269 & 1.286 & 1.250 & 1.230 & \textbf{1.201} & 1.235 & 1.310 & 1.233 & 1.248 & 1.274 & 1.298 & 1.278 & 1.254 \\
 & \textbf{VR} & 0.113 & \textbf{0.000} & 0.155 & 0.078 & 0.101 & 0.044 & 0.076 & 0.102 & 0.124 & 0.141 & 0.110 & 0.117 & 0.084 & 0.088 \\
 & \textbf{VS} & 0.059 & \textbf{0.000} & 0.081 & 0.054 & 0.052 & 0.049 & 0.059 & 0.074 & 0.064 & 0.073 & 0.065 & 0.074 & 0.058 & 0.058 \\
\bottomrule
\end{tabular}
}
\end{table*}

\begin{table*}[htbp]
\centering
\caption{\textbf{Uncertainty Quantification.} Evaluation using probabilistic metrics (PICP, MPIW, Win. Score) to assess the reliability of safety margins under the Best Fit Decreasing (BFD) algorithm. DLinear is excluded due to the difficulty in generating probabilistic outputs. Best results are highlighted in bold.}
\vspace{-1 em}
\label{tab:risk-metrics-bfd}
\resizebox{\textwidth}{!}{
\begin{tabular}{l|l|ccc cccc ccccccc}
\toprule
\multirow{2}{*}{\textbf{Dataset}} & \multirow{2}{*}{\textbf{Metric}} & \multicolumn{3}{c}{\textbf{Statistical Model}} & \multicolumn{4}{c}{\textbf{Deep Learning Model}} & \multicolumn{7}{c}{\textbf{Foundation Model}} \\
\cmidrule(lr){3-5} \cmidrule(lr){6-9} \cmidrule(lr){10-16}
 & & \textbf{AutoARIMA} & \textbf{AutoETS} & \textbf{AutoTheta} & \textbf{DLinear} & \textbf{PatchTST} & \textbf{DeepAR} & \textbf{TFT} & \textbf{Moirai 2} & \textbf{Chronos 2} & \textbf{TimesFM 2.5} & \textbf{Sundial} & \textbf{TOTO} & \textbf{FlowState-9.1M} & \textbf{Kairos 50m} \\
\midrule
\multirow{3}{*}{\textbf{Huawei2025}}
 & \textbf{PICP} & 0.929 & 0.917 & 0.930 & - & 0.893 & 0.875 & 0.886 & 0.919 & 0.922 & 0.912 & 0.578 & \textbf{0.984} & 0.194 & 0.783 \\
 & \textbf{MPIW} & 0.377 & 0.476 & 0.448 & - & 0.376 & 0.334 & 0.341 & 0.376 & 0.323 & 0.339 & \textbf{0.143} & 1.308 & 0.222 & 0.234 \\
 & \textbf{Win. Score} & 0.643 & 0.791 & \textbf{0.637} & - & 0.814 & 1.014 & 0.905 & 0.639 & 0.639 & 0.711 & 1.251 & 1.351 & 6.521 & 0.903 \\
\midrule
\multirow{3}{*}{\textbf{Azure2019}}
 & \textbf{PICP} & 0.939 & 0.956 & 0.954 & - & 0.962 & 0.781 & 0.896 & 0.934 & 0.923 & 0.903 & 0.628 & \textbf{0.994} & 0.194 & 0.779 \\
 & \textbf{MPIW} & 0.372 & 0.415 & 0.420 & - & 0.546 & 0.342 & 0.484 & 0.307 & 0.276 & 0.260 & \textbf{0.139} & 1.562 & 0.197 & 0.225 \\
 & \textbf{Win. Score} & 0.586 & 0.531 & 0.534 & - & 0.680 & 0.965 & 0.937 & 0.450 & \textbf{0.446} & 0.514 & 0.902 & 1.574 & 3.543 & 0.808 \\
\midrule
\multirow{3}{*}{\textbf{Borg2019-d}}
 & \textbf{PICP} & 0.881 & 0.866 & 0.877 & - & 0.856 & 0.614 & \textbf{0.929} & 0.926 & 0.924 & 0.923 & 0.593 & 0.925 & 0.456 & 0.741 \\
 & \textbf{MPIW} & 0.749 & 0.809 & 0.727 & - & 0.360 & \textbf{0.179} & 0.474 & 0.559 & 0.465 & 0.486 & 0.223 & 0.629 & 0.305 & 0.327 \\
 & \textbf{Win. Score} & 1.058 & 1.313 & 1.080 & - & 0.611 & 1.423 & 0.579 & 0.687 & \textbf{0.574} & 0.601 & 1.138 & 0.803 & 2.487 & 0.953 \\
\midrule
\multirow{3}{*}{\textbf{Borg2019-e}}
 & \textbf{PICP} & 0.857 & 0.866 & 0.773 & - & 0.875 & 0.847 & 0.797 & \textbf{0.931} & 0.875 & 0.877 & 0.459 & 0.922 & 0.178 & 0.747 \\
 & \textbf{MPIW} & 0.231 & 0.239 & 0.225 & - & 0.203 & 0.156 & 0.169 & 0.248 & 0.191 & 0.199 & \textbf{0.083} & 0.355 & 0.135 & 0.138 \\
 & \textbf{Win. Score} & 0.389 & 0.400 & 0.624 & - & 0.292 & \textbf{0.268} & 0.359 & 0.327 & 0.309 & 0.321 & 0.747 & 0.431 & 2.557 & 0.366 \\
\bottomrule
\end{tabular}
}
\end{table*}

\begin{table*}[htbp]
\centering
\caption{\textbf{Resource Efficiency and Service Reliability.} End-to-end performance metrics (Util, PAR, VR, VS) evaluated using the Ant Colony Optimization (ACO) algorithm. Best results are highlighted in bold.}
\vspace{-1 em}
\label{tab:efficiency-metrics-aco}
\resizebox{\textwidth}{!}{
\begin{tabular}{l|l|ccc cccc ccccccc}
\toprule
\multirow{2}{*}{\textbf{Dataset}} & \multirow{2}{*}{\textbf{Metric}} & \multicolumn{3}{c}{\textbf{Statistical Model}} & \multicolumn{4}{c}{\textbf{Deep Learning Model}} & \multicolumn{7}{c}{\textbf{Foundation Model}} \\
\cmidrule(lr){3-5} \cmidrule(lr){6-9} \cmidrule(lr){10-16}
 & & \textbf{AutoARIMA} & \textbf{AutoETS} & \textbf{AutoTheta} & \textbf{DLinear} & \textbf{PatchTST} & \textbf{DeepAR} & \textbf{TFT} & \textbf{Moirai 2} & \textbf{Chronos 2} & \textbf{TimesFM 2.5} & \textbf{Sundial} & \textbf{TOTO} & \textbf{FlowState-9.1M} & \textbf{Kairos 50m} \\
\midrule
\multirow{4}{*}{\textbf{Huawei2025}}
 & \textbf{Util} & 0.948 & 0.960 & 0.960 & 0.957 & 0.937 & \textbf{0.969} & \textbf{0.969} & 0.967 & 0.951 & \textbf{0.969} & 0.960 & 0.960 & 0.955 & 0.966 \\
 & \textbf{PAR} & 1.084 & 1.123 & 1.086 & 1.084 & 1.105 & 1.097 & 1.070 & 1.072 & 1.089 & 1.080 & \textbf{1.067} & 1.097 & 1.089 & 1.081 \\
 & \textbf{VR} & 0.198 & 0.259 & 0.216 & 0.173 & \textbf{0.103} & 0.241 & 0.206 & 0.206 & 0.183 & 0.207 & 0.213 & 0.181 & 0.185 & 0.222 \\
 & \textbf{VS} & 0.014 & 0.018 & 0.015 & 0.017 & \textbf{0.013} & 0.015 & 0.015 & 0.015 & 0.014 & 0.015 & 0.014 & 0.016 & 0.014 & 0.014 \\
\midrule
\multirow{4}{*}{\textbf{Azure2019}}
 & \textbf{Util} & 0.916 & 0.916 & 0.916 & 0.927 & \textbf{0.973} & \textbf{0.973} & \textbf{0.973} & 0.922 & 0.918 & 0.936 & 0.920 & 0.918 & \textbf{0.973} & \textbf{0.973} \\
 & \textbf{PAR} & 1.110 & 1.105 & 1.111 & 1.117 & 1.084 & 1.074 & 1.113 & 1.106 & 1.114 & 1.118 & 1.124 & 1.127 & \textbf{1.064} & 1.072 \\
 & \textbf{VR} & \textbf{0.102} & \textbf{0.102} & \textbf{0.102} & 0.125 & 0.222 & 0.285 & 0.520 & 0.151 & 0.124 & 0.153 & 0.143 & 0.145 & 0.193 & 0.238 \\
 & \textbf{VS} & \textbf{0.017} & 0.018 & 0.018 & 0.020 & 0.021 & 0.019 & 0.027 & 0.020 & 0.018 & 0.020 & 0.019 & 0.020 & 0.020 & 0.020 \\
\midrule
\multirow{4}{*}{\textbf{Borg2019-d}}
 & \textbf{Util} & 0.857 & \textbf{0.918} & 0.861 & 0.880 & 0.847 & 0.815 & 0.883 & 0.893 & 0.857 & 0.876 & 0.900 & 0.880 & 0.853 & 0.897 \\
 & \textbf{PAR} & 1.199 & 1.236 & 1.218 & 1.176 & 1.237 & 1.353 & 1.244 & 1.153 & \textbf{1.143} & 1.177 & \textbf{1.143} & 1.182 & 1.177 & 1.169 \\
 & \textbf{VR} & 0.198 & 0.312 & 0.220 & 0.286 & 0.159 & \textbf{0.119} & 0.326 & 0.270 & 0.169 & 0.295 & 0.312 & 0.223 & 0.196 & 0.309 \\
 & \textbf{VS} & 0.063 & 0.094 & 0.071 & 0.080 & 0.059 & 0.060 & 0.080 & 0.070 & \textbf{0.054} & 0.070 & 0.071 & 0.065 & 0.062 & 0.073 \\
\midrule
\multirow{4}{*}{\textbf{Borg2019-e}}
 & \textbf{Util} & 0.637 & 0.637 & \textbf{0.675} & 0.637 & 0.637 & 0.637 & 0.637 & 0.637 & 0.667 & \textbf{0.675} & 0.637 & 0.645 & 0.637 & 0.637 \\
 & \textbf{PAR} & 1.274 & 1.267 & 1.285 & 1.243 & 1.228 & \textbf{1.182} & 1.228 & 1.311 & 1.237 & 1.239 & 1.281 & 1.301 & 1.285 & 1.249 \\
 & \textbf{VR} & 0.112 & \textbf{0.000} & 0.157 & 0.081 & 0.099 & 0.040 & 0.074 & 0.105 & 0.127 & 0.141 & 0.112 & 0.119 & 0.082 & 0.090 \\
 & \textbf{VS} & 0.057 & \textbf{0.000} & 0.081 & 0.058 & 0.050 & 0.035 & 0.057 & 0.071 & 0.062 & 0.073 & 0.064 & 0.074 & 0.065 & 0.059 \\
\bottomrule
\end{tabular}
}
\end{table*}

\begin{table*}[htbp]
\centering
\caption{\textbf{Uncertainty Quantification.} Evaluation using probabilistic metrics (PICP, MPIW, Win. Score) to assess the reliability of safety margins under the Ant Colony Optimization (ACO) algorithm. DLinear is excluded due to the difficulty in generating probabilistic outputs. Best results are highlighted in bold.}
\vspace{-1 em}
\label{tab:risk-metrics-aco}
\resizebox{\textwidth}{!}{
\begin{tabular}{l|l|ccc cccc ccccccc}
\toprule
\multirow{2}{*}{\textbf{Dataset}} & \multirow{2}{*}{\textbf{Metric}} & \multicolumn{3}{c}{\textbf{Statistical Model}} & \multicolumn{4}{c}{\textbf{Deep Learning Model}} & \multicolumn{7}{c}{\textbf{Foundation Model}} \\
\cmidrule(lr){3-5} \cmidrule(lr){6-9} \cmidrule(lr){10-16}
 & & \textbf{AutoARIMA} & \textbf{AutoETS} & \textbf{AutoTheta} & \textbf{DLinear} & \textbf{PatchTST} & \textbf{DeepAR} & \textbf{TFT} & \textbf{Moirai 2} & \textbf{Chronos 2} & \textbf{TimesFM 2.5} & \textbf{Sundial} & \textbf{TOTO} & \textbf{FlowState-9.1M} & \textbf{Kairos 50m} \\
\midrule
\multirow{3}{*}{\textbf{Huawei2025}}
 & \textbf{PICP} & 0.929 & 0.916 & 0.929 & - & 0.886 & 0.879 & 0.891 & 0.920 & 0.922 & 0.914 & 0.575 & \textbf{0.984} & 0.195 & 0.784 \\
 & \textbf{MPIW} & 0.378 & 0.475 & 0.449 & - & 0.373 & 0.341 & 0.346 & 0.376 & 0.325 & 0.339 & \textbf{0.143} & 1.307 & 0.222 & 0.234 \\
 & \textbf{Win. Score} & 0.645 & 0.789 & \textbf{0.642} & - & 0.840 & 1.022 & 0.899 & 0.650 & \textbf{0.642} & 0.708 & 1.259 & 1.352 & 6.530 & 0.908 \\
\midrule
\multirow{3}{*}{\textbf{Azure2019}}
 & \textbf{PICP} & 0.942 & 0.956 & 0.954 & - & 0.954 & 0.792 & 0.895 & 0.932 & 0.923 & 0.904 & 0.621 & \textbf{0.994} & 0.188 & 0.778 \\
 & \textbf{MPIW} & 0.372 & 0.415 & 0.420 & - & 0.531 & 0.346 & 0.486 & 0.306 & 0.277 & 0.265 & \textbf{0.139} & 1.562 & 0.197 & 0.225 \\
 & \textbf{Win. Score} & 0.576 & 0.529 & 0.539 & - & 0.684 & 0.948 & 0.940 & 0.451 & \textbf{0.449} & 0.531 & 0.928 & 1.572 & 3.535 & 0.792 \\
\midrule
\multirow{3}{*}{\textbf{Borg2019-d}}
 & \textbf{PICP} & 0.882 & 0.870 & 0.874 & - & 0.860 & 0.615 & \textbf{0.944} & 0.929 & 0.930 & 0.922 & 0.595 & 0.925 & 0.453 & 0.742 \\
 & \textbf{MPIW} & 0.755 & 0.816 & 0.729 & - & 0.374 & \textbf{0.181} & 0.482 & 0.564 & 0.466 & 0.491 & 0.225 & 0.631 & 0.304 & 0.327 \\
 & \textbf{Win. Score} & 1.078 & 1.310 & 1.078 & - & 0.599 & 1.405 & \textbf{0.554} & 0.682 & 0.566 & 0.609 & 1.141 & 0.805 & 2.480 & 0.960 \\
\midrule
\multirow{3}{*}{\textbf{Borg2019-e}}
 & \textbf{PICP} & 0.864 & 0.862 & 0.777 & - & 0.895 & 0.844 & 0.813 & \textbf{0.929} & 0.872 & 0.879 & 0.462 & 0.918 & 0.186 & 0.747 \\
 & \textbf{MPIW} & 0.232 & 0.239 & 0.225 & - & 0.204 & 0.157 & 0.172 & 0.249 & 0.191 & 0.200 & \textbf{0.083} & 0.355 & 0.136 & 0.137 \\
 & \textbf{Win. Score} & 0.387 & 0.408 & 0.618 & - & 0.280 & \textbf{0.259} & 0.327 & 0.331 & 0.303 & 0.321 & 0.745 & 0.430 & 2.551 & 0.362 \\
\bottomrule
\end{tabular}
}
\end{table*}

\begin{table*}[htbp]
\centering
\caption{\textbf{Resource Efficiency and Service Reliability.} End-to-end performance metrics (Util, PAR, VR, VS) evaluated using the Gurobi optimizer. Best results are highlighted in bold.}
\vspace{-1 em}
\label{tab:efficiency-metrics-gurobi}
\resizebox{\textwidth}{!}{
\begin{tabular}{l|l|ccc cccc ccccccc}
\toprule
\multirow{2}{*}{\textbf{Dataset}} & \multirow{2}{*}{\textbf{Metric}} & \multicolumn{3}{c}{\textbf{Statistical Model}} & \multicolumn{4}{c}{\textbf{Deep Learning Model}} & \multicolumn{7}{c}{\textbf{Foundation Model}} \\
\cmidrule(lr){3-5} \cmidrule(lr){6-9} \cmidrule(lr){10-16}
 & & \textbf{AutoARIMA} & \textbf{AutoETS} & \textbf{AutoTheta} & \textbf{DLinear} & \textbf{PatchTST} & \textbf{DeepAR} & \textbf{TFT} & \textbf{Moirai 2} & \textbf{Chronos 2} & \textbf{TimesFM 2.5} & \textbf{Sundial} & \textbf{TOTO} & \textbf{FlowState-9.1M} & \textbf{Kairos 50m} \\
\midrule
\multirow{4}{*}{\textbf{Huawei2025}}
 & \textbf{Util} & 0.967 & \textbf{0.973} & 0.969 & 0.962 & 0.946 & 0.969 & 0.969 & 0.969 & 0.969 & 0.969 & 0.969 & 0.964 & 0.969 & 0.969 \\
 & \textbf{PAR} & 1.049 & 1.065 & 1.056 & 1.055 & 1.067 & 1.077 & 1.060 & 1.049 & 1.050 & 1.055 & \textbf{1.040} & 1.050 & 1.047 & 1.053 \\
 & \textbf{VR} & 0.189 & 0.253 & 0.208 & 0.152 & \textbf{0.073} & 0.208 & 0.184 & 0.201 & 0.192 & 0.192 & 0.196 & 0.172 & 0.187 & 0.211 \\
 & \textbf{VS} & 0.014 & 0.016 & 0.015 & 0.015 & \textbf{0.013} & 0.016 & 0.015 & 0.014 & \textbf{0.013} & 0.015 & 0.013 & 0.014 & 0.014 & 0.014 \\
\midrule
\multirow{4}{*}{\textbf{Azure2019}}
 & \textbf{Util} & 0.961 & \textbf{0.973} & 0.968 & \textbf{0.973} & \textbf{0.973} & \textbf{0.973} & \textbf{0.973} & \textbf{0.973} & \textbf{0.973} & \textbf{0.973} & \textbf{0.973} & \textbf{0.973} & \textbf{0.973} & \textbf{0.973} \\
 & \textbf{PAR} & 1.061 & 1.063 & \textbf{1.056} & 1.077 & 1.078 & 1.087 & 1.105 & 1.062 & 1.066 & 1.081 & 1.065 & 1.075 & 1.074 & 1.088 \\
 & \textbf{VR} & \textbf{0.119} & 0.190 & 0.157 & 0.152 & 0.203 & 0.247 & 0.331 & 0.197 & 0.176 & 0.192 & 0.174 & 0.193 & 0.199 & 0.230 \\
 & \textbf{VS} & \textbf{0.017} & 0.020 & 0.018 & 0.021 & 0.023 & 0.022 & 0.027 & 0.019 & 0.019 & 0.021 & 0.020 & 0.020 & 0.021 & 0.022 \\
\midrule
\multirow{4}{*}{\textbf{Borg2019-d}}
 & \textbf{Util} & 0.886 & 0.922 & 0.890 & 0.935 & 0.863 & 0.850 & 0.890 & 0.914 & 0.893 & 0.921 & 0.918 & 0.900 & 0.897 & \textbf{0.944} \\
 & \textbf{PAR} & 1.124 & 1.183 & 1.142 & 1.171 & 1.233 & 1.313 & 1.218 & 1.144 & \textbf{1.096} & 1.131 & 1.123 & 1.124 & 1.146 & 1.125 \\
 & \textbf{VR} & 0.173 & 0.266 & 0.205 & 0.290 & 0.122 & \textbf{0.110} & 0.231 & 0.246 & 0.167 & 0.258 & 0.275 & 0.204 & 0.183 & 0.310 \\
 & \textbf{VS} & 0.050 & 0.073 & 0.059 & 0.068 & 0.053 & 0.058 & 0.066 & 0.064 & \textbf{0.045} & 0.061 & 0.061 & 0.054 & 0.054 & 0.068 \\
\midrule
\multirow{4}{*}{\textbf{Borg2019-e}}
 & \textbf{Util} & 0.645 & 0.637 & \textbf{0.691} & 0.637 & 0.637 & 0.637 & 0.637 & 0.637 & 0.667 & 0.675 & 0.637 & 0.645 & 0.637 & 0.637 \\
 & \textbf{PAR} & 1.228 & 1.204 & 1.213 & 1.187 & 1.191 & \textbf{1.133} & 1.184 & 1.226 & 1.185 & 1.184 & 1.225 & 1.229 & 1.185 & 1.212 \\
 & \textbf{VR} & 0.052 & 0.038 & \textbf{0.000} & 0.033 & \textbf{0.000} & \textbf{0.000} & 0.030 & 0.043 & 0.064 & 0.068 & 0.046 & \textbf{0.000} & \textbf{0.000} & 0.035 \\
 & \textbf{VS} & 0.068 & 0.046 & \textbf{0.000} & 0.048 & \textbf{0.000} & \textbf{0.000} & 0.038 & 0.058 & 0.061 & 0.068 & 0.059 & \textbf{0.000} & \textbf{0.000} & 0.061 \\
\bottomrule
\end{tabular}
}
\end{table*}

\begin{table*}[htbp]
\centering
\caption{\textbf{Uncertainty Quantification.} Evaluation using probabilistic metrics (PICP, MPIW, Win. Score) to assess the reliability of safety margins under the Gurobi optimizer. DLinear is excluded due to the difficulty in generating probabilistic outputs. Best results are highlighted in bold.}
\vspace{-1 em}
\label{tab:risk-metrics-gurobi}
\resizebox{\textwidth}{!}{
\begin{tabular}{l|l|ccc cccc ccccccc}
\toprule
\multirow{2}{*}{\textbf{Dataset}} & \multirow{2}{*}{\textbf{Metric}} & \multicolumn{3}{c}{\textbf{Statistical Model}} & \multicolumn{4}{c}{\textbf{Deep Learning Model}} & \multicolumn{7}{c}{\textbf{Foundation Model}} \\
\cmidrule(lr){3-5} \cmidrule(lr){6-9} \cmidrule(lr){10-16}
 & & \textbf{AutoARIMA} & \textbf{AutoETS} & \textbf{AutoTheta} & \textbf{DLinear} & \textbf{PatchTST} & \textbf{DeepAR} & \textbf{TFT} & \textbf{Moirai 2} & \textbf{Chronos 2} & \textbf{TimesFM 2.5} & \textbf{Sundial} & \textbf{TOTO} & \textbf{FlowState-9.1M} & \textbf{Kairos 50m} \\
\midrule
\multirow{3}{*}{\textbf{Huawei2025}}
 & \textbf{PICP} & 0.937 & 0.932 & 0.942 & - & 0.901 & 0.899 & 0.912 & 0.937 & 0.935 & 0.932 & 0.587 & \textbf{0.993} & 0.087 & 0.798 \\
 & \textbf{MPIW} & 0.403 & 0.502 & 0.472 & - & 0.394 & 0.361 & 0.366 & 0.398 & 0.349 & 0.362 & \textbf{0.152} & 1.319 & 0.237 & 0.247 \\
 & \textbf{Win. Score} & 0.670 & 0.744 & 0.660 & - & 0.750 & 0.836 & 0.803 & 0.651 & \textbf{0.637} & 0.677 & 1.226 & 1.340 & 6.106 & 0.847 \\
\midrule
\multirow{3}{*}{\textbf{Azure2019}}
 & \textbf{PICP} & 0.944 & 0.956 & 0.953 & - & 0.959 & 0.812 & 0.911 & 0.931 & 0.915 & 0.896 & 0.626 & \textbf{0.996} & 0.196 & 0.795 \\
 & \textbf{MPIW} & 0.383 & 0.429 & 0.432 & - & 0.536 & 0.347 & 0.490 & 0.321 & 0.289 & 0.274 & \textbf{0.147} & 1.646 & 0.197 & 0.225 \\
 & \textbf{Win. Score} & 0.557 & 0.535 & 0.551 & - & 0.693 & 0.913 & 0.827 & 0.483 & \textbf{0.475} & 0.539 & 0.931 & 1.656 & 3.553 & 0.728 \\
\midrule
\multirow{3}{*}{\textbf{Borg2019-d}}
 & \textbf{PICP} & 0.950 & 0.922 & 0.944 & - & 0.936 & 0.603 & 0.969 & 0.973 & 0.967 & 0.972 & 0.675 & \textbf{0.974} & 0.479 & 0.794 \\
 & \textbf{MPIW} & 0.804 & 0.835 & 0.773 & - & 0.402 & \textbf{0.189} & 0.512 & 0.593 & 0.507 & 0.537 & 0.240 & 0.657 & 0.330 & 0.349 \\
 & \textbf{Win. Score} & 0.935 & 1.053 & 0.914 & - & \textbf{0.525} & 1.353 & 0.551 & 0.639 & 0.569 & 0.580 & 0.917 & 0.717 & 1.825 & 0.741 \\
\midrule
\multirow{3}{*}{\textbf{Borg2019-e}}
 & \textbf{PICP} & 0.882 & 0.873 & 0.788 & - & 0.925 & 0.850 & 0.863 & 0.936 & 0.902 & 0.884 & 0.488 & \textbf{0.956} & 0.271 & 0.742 \\
 & \textbf{MPIW} & 0.258 & 0.255 & 0.248 & - & 0.222 & 0.164 & 0.184 & 0.275 & 0.207 & 0.216 & \textbf{0.092} & 0.388 & 0.147 & 0.150 \\
 & \textbf{Win. Score} & 0.413 & 0.439 & 0.590 & - & 0.295 & \textbf{0.284} & 0.293 & 0.356 & 0.313 & 0.348 & 0.835 & 0.452 & 2.329 & 0.460 \\
\bottomrule
\end{tabular}
}
\vspace{1em}
\end{table*}

\begin{table*}[!htbp]
\centering
\small
\renewcommand{\arraystretch}{0.88} 
\caption{Comparison of Zero-shot and Fine-tuned Performance of Representative Time Series Foundation Models.}
\vspace{-0.8 em}
\label{tab:finetuned_tsfm}
\begin{tabular}{ll cccc cccc}
\toprule
\multirow{2}{*}{\textbf{Dataset}} & \multirow{2}{*}{\textbf{Model}} & \multicolumn{4}{c}{\textbf{Zero-shot}} & \multicolumn{4}{c}{\textbf{Fine-tuned}} \\
\cmidrule(lr){3-6} \cmidrule(lr){7-10}
& & \textbf{MASE} $\downarrow$ & \textbf{CRPS} $\downarrow$ & \textbf{Util} $\uparrow$ & \textbf{VR} $\downarrow$ & \textbf{MASE} $\downarrow$ & \textbf{CRPS} $\downarrow$ & \textbf{Util} $\uparrow$ & \textbf{VR} $\downarrow$ \\
\midrule

\multirow{3}{*}{\textbf{Huawei2025}} 
& Chronos 2 & 0.752 & 0.642 & 0.948 & 0.183 & 0.742 & 0.664 & 0.948 & 0.186 \\
& Moirai 2  & 0.869 & 0.760 & 0.951 & 0.187 & 0.747 & 0.719 & 0.967 & 0.205 \\
& TOTO      & 0.881 & 0.718 & 0.937 & 0.074 & 0.863 & 0.707 & 0.959 & 0.186 \\
\midrule

\multirow{3}{*}{\textbf{Azure2019}} 
& Chronos 2 & 0.690 & 0.364 & 0.920 & 0.135 & 0.657 & 0.325 & 0.915 & 0.121 \\
& Moirai 2  & 0.713 & 0.364 & 0.915 & 0.121 & 0.677 & 0.333 & 0.926 & 0.153 \\
& TOTO      & 0.706 & 0.361 & 0.915 & 0.122 & 0.660 & 0.327 & 0.917 & 0.150 \\
\midrule

\multirow{3}{*}{\textbf{Borg2019-d}} 
& Chronos 2 & 0.702 & 0.625 & 0.869 & 0.217 & 0.667 & 0.586 & 0.854 & 0.166 \\
& Moirai 2  & 0.774 & 0.717 & 0.890 & 0.276 & 0.717 & 0.626 & 0.891 & 0.280 \\
& TOTO      & 0.852 & 0.758 & 0.869 & 0.218 & 0.724 & 0.665 & 0.879 & 0.231 \\
\midrule

\multirow{3}{*}{\textbf{Borg2019-e}} 
& Chronos 2 & 0.880 & 0.341 & 0.637 & 0.102 & 0.844 & 0.323 & 0.667 & 0.123 \\
& Moirai 2  & 0.852 & 0.336 & 0.637 & 0.092 & 0.847 & 0.341 & 0.637 & 0.101 \\
& TOTO      & 0.935 & 0.365 & 0.637 & 0.124 & 0.859 & 0.330 & 0.644 & 0.117 \\
\bottomrule
\end{tabular}
\end{table*}

This section details the experimental configurations for time series forecasting and resource consolidation simulation.
\vspace{-0.5 em}
\subsection{Time Series Forecasting Settings}
\vspace{-0.2 em}
For the forecasting task evaluated in Section \ref{sec-4.1}, we chronologically split each dataset into training and testing sets using an 80/20 ratio. We employ a non-overlapping rolling window forecasting strategy. 

\noindent \textbf{Window Configurations:} For deep learning and foundation models, we adopt frequency-specific setups for look-back windows and forecast horizons:
\begin{itemize}[leftmargin=*, nosep, noitemsep]
    \tightlist
    \item \textbf{5-minute frequency:} The look-back window is fixed to 512. Forecast horizons include \{48, 96, 192, 336\}.
    \item \textbf{30-minute frequency:} The look-back window is fixed to 96. Forecast horizons include \{12, 24, 36, 48\}.
    \item \textbf{1-hour frequency:} The look-back window is fixed to 36. Forecast horizons include \{6, 8, 10, 12\}.
\end{itemize}
For statistical models, the historical input length is truncated to be less than or equal to 512 to ensure evaluation efficiency.

\noindent \textbf{Training and Hyperparameters:} 
The training and inference processes vary among different families of models. Statistical models make predictions by directly analyzing historical data patterns and do not require a separate training phase. Deep learning models need to be trained separately on each dataset for different experimental settings. We perform hyperparameter tuning using Optuna \cite{optuna} across 15 runs for each model. The hyperparameter search range for model specific hyperparameters is given in Table \ref{tab:hyperparameters}, and defined as such by picking values surrounding the default values provided by their respective papers’ official implementations. On top of the
listed parameters for each model, we also search for weight decay on all runs in the range: $[ 10^{-8}, 10^{-5}]$. Additionally, we set the batch size to 64 and, number of batches per epoch to 100, and finally number of epochs to 100. For the foundation models, our primary focus is on their zero-shot capabilities. Therefore, no specialized fine-tuning is performed.

\vspace{-0.5 em}
\subsection{Resource Consolidation Simulation Settings}
\vspace{-0.2 em}
To ensure a rigorous and reproducible evaluation for the experiments in Sections \ref{sec-4.2} and \ref{sec-4.3}, we implement the simulation environment using SimPy \cite{simpy}, a process-based discrete-event simulation framework. Experiments focus on the hourly-granularity subset of the multi-granularity datasets to balance simulation fidelity with computational efficiency, while sufficiently capturing diurnal workload patterns.

\noindent \textbf{Data Preparation and Simulation Scale:} 
We apply a rigorous preprocessing pipeline to construct the evaluation scenarios.
\begin{itemize}[leftmargin=*, nosep, noitemsep]
    \item Workload Aggregation: For Borg2019, original task-level traces are aggregated into job-level workloads to represent logical service units.
    \item Filtering: Traces shorter than 336 time steps (14 days) or exhibiting extreme average CPU utilization ($< 1\%$ or $> 99\%$) are excluded.
    \item Downsampling and Scale: Datasets exceeding 1,000 samples are downsampled with a ratio of 0.1 to balance computational efficiency. The final simulation scale for each dataset is as follows: Huawei2025 (575 VMs), Azure2019 (150 VMs), Borg2019-d (127 VMs), and Borg2019-e (68 VMs).
\end{itemize}

\noindent \textbf{Physical Machine Configuration:}
We simulate a homogeneous cluster environment where all PMs possess identical resource specifications. The CPU capacity of each PM is normalized. Consistent with previous works \cite{luo2021correlation,luo2021intelligent,vm-review,con-survey1}, we focus on CPU utilization as the primary dimension for resource consolidation, as it represents the predominant bottleneck in data center energy consumption.

\noindent \textbf{Sliding Window Strategy:} 
We conduct simulation experiments on datasets spanning a 14-day period. Evaluation scenarios are generated using a sliding window strategy with a stride of 1. Each scenario encompasses the workload data for all VMs, consisting of their 5-day historical usage and 24-hour future demand sequences. Data spanning Day 1 to Day 12 is utilized for deep learning model training. The final 2 days are reserved for testing, yielding 25 test scenarios per dataset.

\begin{figure*}[!htbp]
    \centering  
    \includegraphics[width=0.95\textwidth]{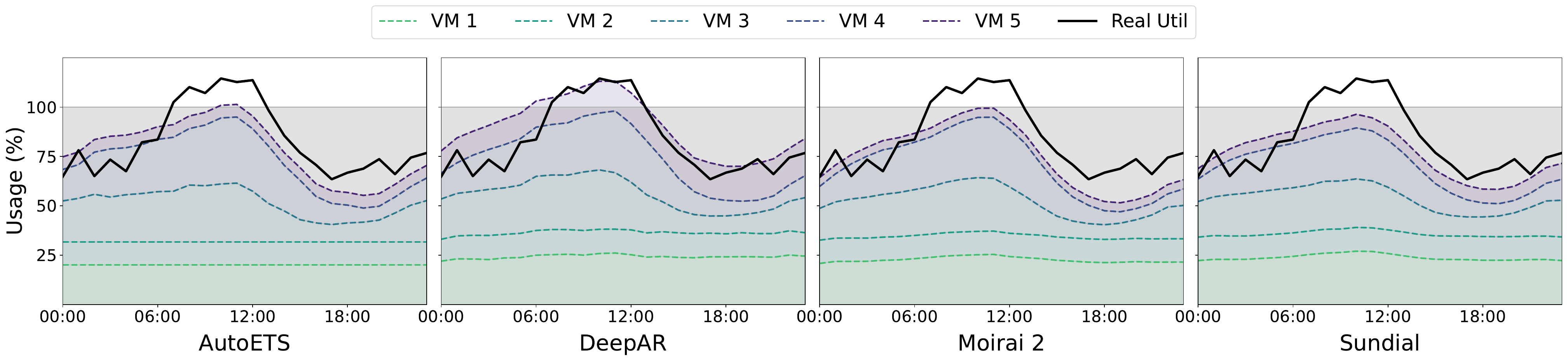}
    \vspace{-0.8 em}
    \caption{Visualization of resource consolidation on a single physical machine (PM) using different forecasting models. Stacked colored dashed lines represent the predicted future demands of individual VMs allocated to the PM. The black solid line denotes the actual aggregated resource usage of the PM, while the grey shaded region indicates the PM's capacity limit.}
    \label{fig:case0}
\end{figure*}
\section{Supplementary Experiments and Evaluation}

\subsection{Detailed Evaluation Results per Optimization Method}
\label{sec:appendix_detailed_results}

To provide a more detailed view of the CloudCons benchmark, this section presents the disaggregated experimental results for each optimization method: First Fit Decreasing (FFD), Best Fit Decreasing (BFD), Ant Colony Optimization (ACO), and Gurobi. 

Specifically, Tables \ref{tab:efficiency-metrics-ffd}--\ref{tab:risk-metrics-gurobi} report the comprehensive performance under each optimization method. These tables detail both the resource efficiency and service reliability metrics (Util, PAR, VR, VS) to evaluate the end-to-end decision utility, and the uncertainty quantification metrics (PICP, MPIW, Win. Score) to assess the reliability of safety margins.

\subsection{Fine-Tuning Experiments for Foundation Models}
\label{sec:tsfm-finetune}

To further explore the potential of time series foundation models in the specific context of cloud resource consolidation, we conduct supplemental fine-tuning experiments on representative models. As presented in Table \ref{tab:finetuned_tsfm}, fine-tuning expectedly improves forecasting accuracy across most datasets. However, this predictive enhancement does not guarantee corresponding gains in downstream decision utility. For instance, when evaluating TOTO on the Azure2019 dataset, fine-tuning reduces MASE and CRPS; however, while resource utilization remains virtually unchanged, the violation rate paradoxically deteriorates from 0.122 to 0.150. A similar degradation in service reliability is observed when testing Chronos 2 on the Huawei2025 dataset. These empirical results further substantiate our core conclusion: superior forecasting accuracy does not inherently translate into better downstream resource consolidation outcomes.

\section{Case Study: The Cause of Service Violations}
\label{case-study}

To illustrate how forecasting deviations directly influence downstream decision-making, we present a representative consolidation case from the Borg2019-e dataset, which is distinctively characterized by highly synchronized 24-hour diurnal rhythms. In this scenario, five VMs are allocated to a single PM based on the consolidation decision derived from Moirai 2’s forecasts. For comparison, we also employ AutoETS, DeepAR, and Sundial to predict the future demands of these VMs. Figure \ref{fig:case0} visualizes the predicted demands (colored dashed curves), where the cumulative height represents the total predicted load on the PM, compared against the actual aggregated resource usage (black solid line). As observed, both Moirai 2 and Sundial significantly underestimate the aggregated demand peak, misleading the optimizer into identifying this assignment as a feasible and safe decision. In contrast, AutoETS predicts a peak that marginally exceeds the capacity, signaling potential risk despite not fully capturing the magnitude of the actual usage. Meanwhile, DeepAR demonstrates superior fidelity, closely tracking the actual high-load behavior with predictions. Consequently, using these conservative forecasts, particularly DeepAR, would have allowed this risky consolidation to be correctly identified as a capacity violation and rejected. This highlights the misalignment: despite Moirai 2 and Sundial achieving superior forecasting accuracy on this dataset, their failure to capture peaks compromises decision utility, whereas the baselines with higher overall prediction errors effectively avert severe resource contention.

\end{document}